%% file: kmodes.tex
\documentclass[letterpaper]{article}
\usepackage[margin=1in,dvips]{geometry}
\usepackage{graphicx,psfrag,amssymb,amsthm,amsmath}
\usepackage{natbib}
\usepackage{url,color}

\input{MACPdef}
\input{MACPdef-ams}

\graphicspath{{grf/}}

\bibpunct[, ]{(}{)}{;}{a}{,}{,} 

\begin{document}

\title{The $K$-Modes Algorithm for Clustering}
\author{
  Miguel {\'A}.\ Carreira-Perpi{\~n}{\'a}n \hspace{5ex} Weiran Wang \\
  Electrical Engineering and Computer Science, University of California, Merced \\
  {\small\url{http://eecs.ucmerced.edu}}
}
\date{April 23, 2013}

\AtBeginDvi{}

\maketitle

\begin{abstract}
  Many clustering algorithms exist that estimate a cluster centroid, such as $K$-means, $K$-medoids or mean-shift, but no algorithm seems to exist that clusters data by returning exactly $K$ meaningful modes. We propose a natural definition of a $K$-modes objective function by combining the notions of density and cluster assignment. The algorithm becomes $K$-means and $K$-medoids in the limit of very large and very small scales. Computationally, it is slightly slower than $K$-means but much faster than mean-shift or $K$-medoids. Unlike $K$-means, it is able to find centroids that are valid patterns, truly representative of a cluster, even with nonconvex clusters, and appears robust to outliers and misspecification of the scale and number of clusters.
\end{abstract}


Given a dataset $\x_1,\dots,\x_N \in \bbR^D$, we consider clustering algorithms based on centroids, i.e., that estimate a representative $\c_k \in \bbR^D$ of each cluster $k$ in addition to assigning data points to clusters. Two of the most widely used algorithms of this type are $K$-means and mean-shift. $K$-means has the number of clusters $K$ as a user parameter and tries to minimize the objective function
\begin{gather}
  \label{e:kmeans-objfcn}
  \min_{\RR,\C} E(\RR,\C) = \sum^K_{k=1}{\sum^N_{n=1}{r_{nk} \norm{\x_n-\c_k}^2}} \\
  \text{s.t. } r_{nk} \in\{0,1\},\ \sum^K_{k=1}{r_{nk}} = 1,\ n=1,\dots,N,\ k=1,\dots,K\notag
\end{gather}
where \RR\ are binary assignment variables (of point $n$ to cluster $k$) and $\C = (\c_1,\dots,\c_K)$ are centroids, free to move in $\bbR^D$. At an optimum, centroid $\c_k$ is the mean of the points in its cluster. Gaussian mean-shift \cite{FukunagHostet75a,Cheng95a,Carreir00b,ComanicMeer02a} assumes we have a kernel density estimate (kde) with bandwidth $\sigma > 0$ and kernel $G(t) \propto e^{-t/2}$
\begin{equation}
  \label{e:kde}
  p(\x) = \frac{1}{N} \sum^N_{n=1}{G\bigl(\norm{(\x-\x_n)/\sigma}^2\bigr)} \qquad \x\in\bbR^D
\end{equation}
and applies the iteration (started from each data point):
\begin{gather}
  \label{e:GMS}
  p(n|\x) = \frac{\exp{\bigl(-\frac{1}{2} \norm{(\x-\x_n)/\sigma}^2\bigr)}}{\sum^N_{n'=1}{\exp{\bigl(-\frac{1}{2} \norm{(\x-\x_{n'})/\sigma}^2\bigr)}}}, \qquad \x \leftarrow \f(\x) = \sum^N_{n=1}{p(n|\x) \x_n}.
\end{gather}
which converges to a mode (local maximum) of $p$ from nearly any initial \x\ \cite{Carreir07a}. Each mode is the centroid for one cluster, which contains all the points that converge to its mode. The user parameter is the bandwidth $\sigma$ and the resulting number of clusters depends on it implicitly.

The pros and cons of both algorithms are well known. $K$-means tends to define round clusters; mean-shift can obtain clusters of arbitrary shapes and has been very popular in low-dimensional clustering applications such as image segmentation \cite{ComanicMeer02a}, but does not work well in high dimension. Both can be seen as special EM algorithms \cite{Bishop06a,Carreir07a}. Both suffer from outliers, which can move centroids outside their cluster in $K$-means or create singleton modes in mean-shift. Computationally, $K$-means is much faster than mean-shift, at $\calO(KND)$ and $\calO(N^2D)$ per iteration, respectively, particularly with large datasets. In fact, accelerating mean-shift has been a topic of active research \cite{Carreir06a,Yuan_10a}. Mean-shift does not require a value of $K$, which is sometimes convenient, although many users often find it desirable to force an algorithm to produce exactly $K$ clusters (e.g.\ if prior information is available).

\begin{figure}[b!]
  \begin{center}
    \begin{tabular}{@{}c@{\hspace{0.0033\linewidth}}c@{\hspace{0.0033\linewidth}}c@{\hspace{0.0033\linewidth}}c@{\hspace{0.0033\linewidth}}c@{\hspace{0.0033\linewidth}}c@{\hspace{0.0033\linewidth}}c@{}}
      \includegraphics[width=0.14\linewidth,bb=275 367 327 419,clip]{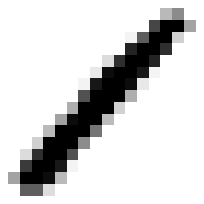} &
      \includegraphics[width=0.14\linewidth,bb=275 367 327 419,clip]{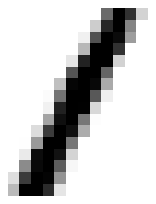} &
      \includegraphics[width=0.14\linewidth,bb=275 367 327 419,clip]{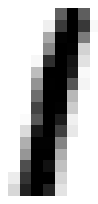} &
      \includegraphics[width=0.14\linewidth,bb=275 367 327 419,clip]{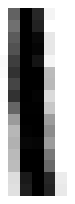} &
      \includegraphics[width=0.14\linewidth,bb=275 367 327 419,clip]{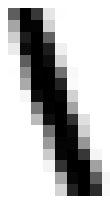} &
      \includegraphics[width=0.14\linewidth,bb=275 367 327 419,clip]{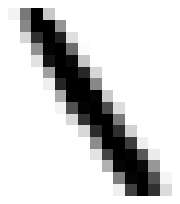} &
      \includegraphics[width=0.14\linewidth,bb=275 367 327 419,clip]{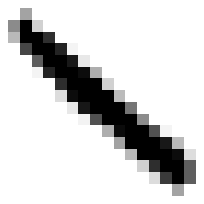}
    \end{tabular} \\                          
    \begin{tabular}{@{}c@{\hspace{0.05\linewidth}}c@{\hspace{0.05\linewidth}}c@{}}
      $K$-means & $K$-modes & GMS \\
      \includegraphics[width=0.23\linewidth,bb=275 367 327 419,clip]{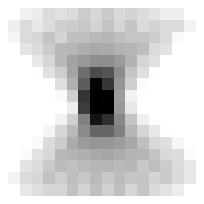} & 
      \includegraphics[width=0.23\linewidth,bb=275 367 327 419,clip]{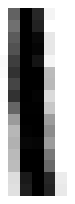} & 
      \includegraphics[width=0.23\linewidth,bb=275 367 327 419,clip]{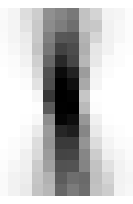} 
    \end{tabular}
    \caption{A cluster of 7 rotated-1 USPS digit images and the centroids found by $K$-means, $K$-modes (both with $K=1$) and mean-shift (with $\sigma$ so there is one mode).}
    \label{f:rotated1}
  \end{center}
\end{figure}

One important aspect in many applications concerns the validity of the centroids as patterns in the input space, as well as how representative they are of their cluster. Fig.~\ref{f:rotated1} illustrates this with a single cluster consisting of continuously rotated digit-1 images. Since these images represent a nonconvex cluster in the high-dimensional pixel space, their mean (which averages all orientations) is not a valid digit-1 image, which makes the centroid not interpretable and hardly representative of a digit 1. Mean-shift does not work well either: to produce a single mode, a large bandwidth is required, which makes the mode lie far from the manifold; a smaller bandwidth does produce valid digit-1 images, but then multiple modes arise for the same cluster, and under mean-shift they define each a cluster. Clustering applications that require valid centroids for nonconvex or manifold data abound (e.g.\ images, shapes or proteins).

A third type of centroid-based algorithms are exemplar-based or $K$-medoid clustering \cite{KaufmanRousseeuw90a,Bishop06a,Hastie_09a}. These constrain the centroids to be points from the dataset (``exemplars''), such as $K$-medians, and often minimize a $K$-means objective function~\eqref{e:kmeans-objfcn} with a non-Euclidean distance. They are slow, since updating centroid $\c_k$ requires testing all pairs of points in cluster $k$. Forcing the centroids to be exemplars is often regarded as a way to ensure the centroids are valid patterns. However, the exemplars themselves are often noisy and thus not that representative of their neighborhood. Not constraining a centroid to be an exemplar can remove such noise and produce a more typical representative.

Given that most location statistics have been used for clustering (mean, mode, median), it is remarkable that no $K$-modes formulation for clustering seems to exist, that is, an algorithm that will find exactly $K$ modes that correspond to meaningful clusters. An obvious way to define a $K$-modes algorithm is to pick $K$ modes from a kde, but it is not clear what modes to pick (assuming it has at least $K$ modes, which will require a sufficiently small bandwidth). Picking the modes with highest density need not correlate well with clusters that have an irregular density, or an approximately uniform density with close but distinct high-density modes.

We define a $K$-modes objective as a natural combination of two ideas: the cluster assignment idea from $K$-means and the density maximization idea of mean-shift. The algorithm has two interesting special cases, $K$-means and a version of $K$-medoids, in the limits of large and small bandwidth, respectively. For small enough bandwidth, the centroids are denoised, valid patterns and typical representatives of their cluster. Computationally, it is slightly slower than $K$-means but much faster than mean-shift or $K$-medoids.

\section{A $K$-modes Objective Function}
\label{s:objfcn}

We maximize the objective function
\begin{gather}
  \label{e:kmodes-objfcn}
  \max_{\RR,\C} L(\RR,\C) = \sum^K_{k=1}{\sum^N_{n=1}{r_{nk} G\bigl(\norm{(\x_n-\c_k)/\sigma}^2\bigr)}} \\
  \text{s.t. } r_{nk} \in\{0,1\},\ \sum^K_{k=1}{r_{nk}} = 1,\ n=1,\dots,N,\ k=1,\dots,K.\notag
\end{gather}
For a given assignment \RR, this can be seen as (proportional to) the sum of a kde as in eq.~\eqref{e:kde} but separately for each cluster. Thus, a good clustering must move centroids to local modes, but also define $K$ separate kdes. This naturally combines the idea of clustering through binary assignment variables with the idea that high-density points are representative of a cluster (for suitable bandwidth values).

As a function of the bandwidth $\sigma$, the $K$-modes objective function has two interesting limit cases. When $\sigma\rightarrow\infty$, it becomes $K$-means. This can be seen from the centroid update (which becomes the mean), or from the objective function directly. Indeed, approximating it with Taylor's theorem for very large $\sigma$ and using the fact that $\smash{\sum^K_{k=1}{r_{nk}} = 1}$ gives
\begin{equation*}
  L(\RR,\C) \approx \sum^N_{k=1}{\sum^N_{n=1}{r_{kn} (1-\norm{\x_n-\c_k}^2/2\sigma^2)}} = N - E(\RR,\C)/2\sigma^2
\end{equation*}
where $E(\RR,\C)$ is the same as in eq.~\eqref{e:kmeans-objfcn} and is subject to the same constraints. Thus, maximizing $L$ becomes minimizing $E$, exactly the $K$-means problem. When $\sigma\rightarrow 0$, it becomes a $K$-medoids algorithm, since the centroids are driven towards data points. Thus, $K$-modes interpolates smoothly between these two algorithms, creating a continuous path that links a $K$-mean to a $K$-medoid. However, its most interesting behavior is for intermediate $\sigma$.

\section{Two $K$-modes Algorithms}
\label{s:alg}

As is the case for $K$-means and $K$-medoids, minimizing the $K$-modes objective function is NP-hard. We focus on iterative algorithms that find a locally optimum clustering in the sense that no improvement is possible on the centroids given the current assignments, and vice versa. We give first an algorithm for fixed $\sigma$ and then use it to construct a homotopy algorithm that sweeps over a $\sigma$ interval.

\subsubsection*{For Fixed $\sigma$}

It is convenient to use alternating optimization:
\begin{description}
\item[Assignment step] Over assignments \RR\ for fixed \C, the constrained problem separates into a constrained problem for each point $\x_n$, of the form
  \begin{equation*}
    \max_{\RR_n}{\sum^K_{k=1}{r_{nk} g_{nk}}} \quad \text{ s.t. } \quad \sum^K_{k=1}{r_{nk}} = 1,\ r_{nk} \in\{0,1\},\ k=1,\dots,K,
  \end{equation*}
with $g_{nk} = G\bigl(\smash{\norm{(\x_n-\c_k)/\sigma}}^2\bigr)$. The solution is given by assigning point $\x_n$ to its closest centroid in Euclidean distance (assuming the kernel $G$ is a decreasing function of the Euclidean distance).
\item[Mode-finding step] Over centroids \C\ for fixed \RR, we have a separate unconstrained maximization for each centroid, of the form
  \begin{equation*}
    L(\c_k) = \sum^N_{n=1}{r_{nk} G\bigl(\norm{(\x_n-\c_k)/\sigma}^2\bigr)},
  \end{equation*}
which is proportional to the cluster kde, and can be done with mean-shift. Note the step over \C\ need not be exact, i.e., the centroids need not converge to their corresponding modes. We exit when a tolerance is met or $I$ mean-shift iterations have been run.
\end{description}
Thus, the algorithm operates similarly to $K$-means but finding modes instead of means: it interleaves a hard assignment step of data points to centroids with a mode-finding step that moves each centroid to a mode of the kde defined by the points currently assigned to it.

Convergence of this algorithm (in value) follows from the facts that each step (over \RR\ or over \C) is strictly feasible and decreases the objective or leaves it unchanged, and that the objective function is lower bounded by 0 within the feasible set. Besides, since there is a finite number of assignments, convergence occurs in a finite number of outer-loop steps (as happens with $K$-means) if the step over \C\ is exact and deterministic. By this we mean that for each $\c_k$ we find deterministically a maximum of its objective function (i.e., the mode for $\c_k$ is a deterministic function of \RR). This prevents the possibility that for the same assignment \RR\ we find different modes for a given $\c_k$, which could lead the algorithm to cycle. This condition can be simply achieved by using an optimization algorithm that either has no user parameters (such as step sizes; mean-shift is an example), or has user parameters set to fixed values, and running it to convergence. The $(\RR^*,\C^*)$ convergence point is a local maximum in the sense that $L(\RR^*,\C)$ has a local maximum at $\C=\C^*$ and $L(\RR,\C^*)$ has a global maximum at $\RR=\RR^*$.

The computational cost per outer-loop iteration of this algorithm (setting $I=1$ for simplicity in the mean-shift step) is identical to that of $K$-means: the step over \RR\ is $\calO(KND)$ and the step over \C\ is $\calO(N_1 D + \dots + N_K D) = \calO(ND)$ (where $N_k$ is the number of points currently assigned to $\c_k$), for a total of $\calO(KND)$. And also as in $K$-means, the steps parallelize: over \C, the mean-shift iteration proceeds independently in each cluster; over \RR, each data point can be processed independently.

\subsubsection*{Homotopy Algorithm}

We start with $\sigma=\infty$ (i.e., run $K$-means, possibly several times and picking the best optimum). Then, we gradually decrease $\sigma$ while running $J$ iterations of the fixed-$\sigma$ $K$-modes algorithm for each value of $\sigma$, until we reach a target value $\sigma^*$. This follows an optimum path $(\RR(\sigma),\C(\sigma))$ for $\sigma\in[\sigma^*,\infty)$. In practice, as is well known with homotopy techniques, this tends to find better optima than starting directly at the target value $\sigma^*$. We use this homotopy algorithm in our experiments. Given we have to run $K$-means multiple times to find a good initial optimum (as commonly done in practice), the homotopy does not add much computation. Note that the homotopy makes $K$-modes a deterministic algorithm given the local optimum found by $K$-means.

\subsubsection*{User Parameters}

The basic user parameter of $K$-modes is the desired number of clusters $K$. The target bandwidth $\sigma^*$ in the homotopy is simply used as a scaling device to refine the centroids. We find that representative, valid centroids are obtained for a wide range of intermediate $\sigma$ values. A good target $\sigma^*$ can be obtained with a classical bandwidth selection criterion for kernel density estimates \cite{WandJones94a}, such as the average distance to the $k$th nearest neighbor.

Practically, a user will typically be interested in the $K$ centroids and clusters resulting for the target bandwidth. However, examining the centroid paths $\c_k(\sigma)$ can also be interesting for exploratory analysis of a dataset, as illustrated in our experiments with handwritten digit images.

\section{Relation with Other Algorithms}
\label{s:related}

$K$-modes is most closely related to $K$-means and to Gaussian mean-shift (GMS), since it essentially introduces the kernel density estimate into the $K$-means objective function. This allows $K$-modes to find exactly $K$ true modes in the data (in its mathematical sense, i.e., maxima of the kde for each cluster), while achieving assignments as in $K$-means, and with a fast runtime, thus enjoying some of the best properties from both $K$-means and GMS.

$K$-means and $K$-modes have the same update step for the assignments, but the update step for the centroids is given by setting each centroid to a different location statistic of the points assigned to it: the mean for $K$-means, a mode for $K$-modes. $K$-means and $K$-modes also define the same class of clusters (a Voronoi tessellation, thus convex clusters), while GMS can produce possibly nonconvex, disconnected clusters.

In GMS, the number of clusters equals the number of modes, which depends on the bandwidth $\sigma$. If one wants to obtain exactly $K$ modes, there are two problems. The first one is computational: since $K$ is an implicit, nonlinear function of $\sigma$, finding a $\sigma$ value that produces $K$ modes requires inverting this function. This can be achieved numerically by running mean-shift iterations while tracking $K(\sigma)$ as in scale-space approaches \cite{Collin03b}, but this is very slow. Besides, particularly for high-dimensional data, the kde only achieves $K$ modes for a very narrow (even empty) interval of $\sigma$. The second problem is that even with an optimally tuned bandwidth, a kde will usually create undesirable, spurious modes where data points are sparse (e.g.\ outliers or cluster boundaries), again particularly with high-dimensional data. We avoid this problem in the homotopy version of $K$-modes by starting with large $\sigma$, which tracks important modes. The difference between $K$-modes and GMS is clearly seen in the particular case where we set $K=1$ (as in fig.~\ref{f:rotated1}): $K$-modes runs the mean-shift update initialized from the data mean, so as $\sigma$ decreases, this will tend to find a single, major mode of the kde. However, the kde itself will have many modes, all of which would become clusters under GMS.

The fundamental problem in GMS is equating modes with clusters. The true density of a cluster may well be multimodal to start with. Besides, in practice a kde will tend to be bumpy unless the bandwidth is unreasonably large, because it is by nature a sum of bumpy kernels centered at the data points. This is particularly so with outliers (which create small modes) or in high dimensions. There is no easy way to smooth out a kde (increasing the bandwidth does smooth it, but at the cost of distorting the overall density) or to postprocess the modes to select ``good'' ones. One has to live with the fact that a good kde will often have multiple modes per cluster.

$K$-modes provides one approach to this problem, by separating the roles of cluster assignment and of density. Each cluster has its own kde, which can be multimodal, and the homotopy algorithm tends to select an important mode among these within each cluster. This allows $K$-modes to achieve good results even in high-dimensional problems, where GMS fails.

Computationally, $K$-modes and $K$-means are $\calO(KND)$ per iteration for a dataset of $N$ points in $D$ dimensions. While $K$-modes in its homotopy version will usually take more iterations, this extra runtime is small because in practice one runs $K$-means multiple times from different initializations to achieve a better optimum. GMS is $\calO(N^2D)$ per iteration, which is far slower, particularly with large datasets. The reason is that in GMS the kde involves all $N$ points and one must run mean-shift iterations started from each of the $N$ points. However, in $K$-modes the kde for cluster $k$ involves only the $N_k$ points assigned to it and one must run mean-shift iterations only for the centroid $\c_k$. Much work has addressed approximating GMS so that it runs faster, and some of it could be applied to the mean-shift step in $K$-modes, such as using Newton or sparse EM iterations \cite{Carreir06a}.

In addition to these advantages, our experiments show that $K$-modes can be more robust than $K$-means and GMS with outliers and with misspecification of either $K$ or $\sigma$.

There are two variations of mean-shift that replace the local mean step of eq.~\eqref{e:GMS} with a different statistic: the local (Tukey) median \cite{Shapir_09a} and a medoid defined as any dataset point which minimizes a weighted sum of squared distances \cite{Sheikh_07a}. Both are really medoid algorithms, since they constrain the centroids to be data points, and do not find true modes (maxima of the density). In general, $K$-medoid algorithms such as $K$-centers or $K$-medians are combinatorial problems, typically NP-hard \cite{HochbaumShmoys85a,KaufmanRousseeuw90a,Meyers_04a}. In the limit $\sigma\rightarrow 0$, $K$-modes can be seen as a deterministic annealing approach to a $K$-medoids objective (just as the elastic net \cite{DurbinWillsha87a} is for the traveling salesman problem).

There exists another algorithm called ``$K$-modes'' \cite{Huang98a,Chatur_01a}. This is defined for categorical data and uses the term ``mode'' in a generic sense of ``centroid''. It is unrelated to our algorithm, which is defined for continuous data and uses ``mode'' in its mathematical sense of density maximum.

\section{Experiments}
\label{s:expts}

We compare with $K$-means and Gaussian mean-shift (GMS) clustering. For $K$-means, we run it 20 times with different initializations and pick the one with minimum value of $E$ in eq.~\eqref{e:kmeans-objfcn}. For $K$-modes, we use its homotopy version initialized from the best $K$-means result and finishing at a target bandwidth (whose value is set either by using a kde bandwidth estimation rule or by hand, depending on the experiment).

\subsubsection*{Toy Examples}

\begin{figure}[t]
  \begin{center}
    \begin{tabular}{@{}c@{\hspace{0\linewidth}}c@{}}
      \psfrag{sigma=10}[b][]{$\sigma=\infty$ ($K$-means)}
      \includegraphics[width=0.50\linewidth]{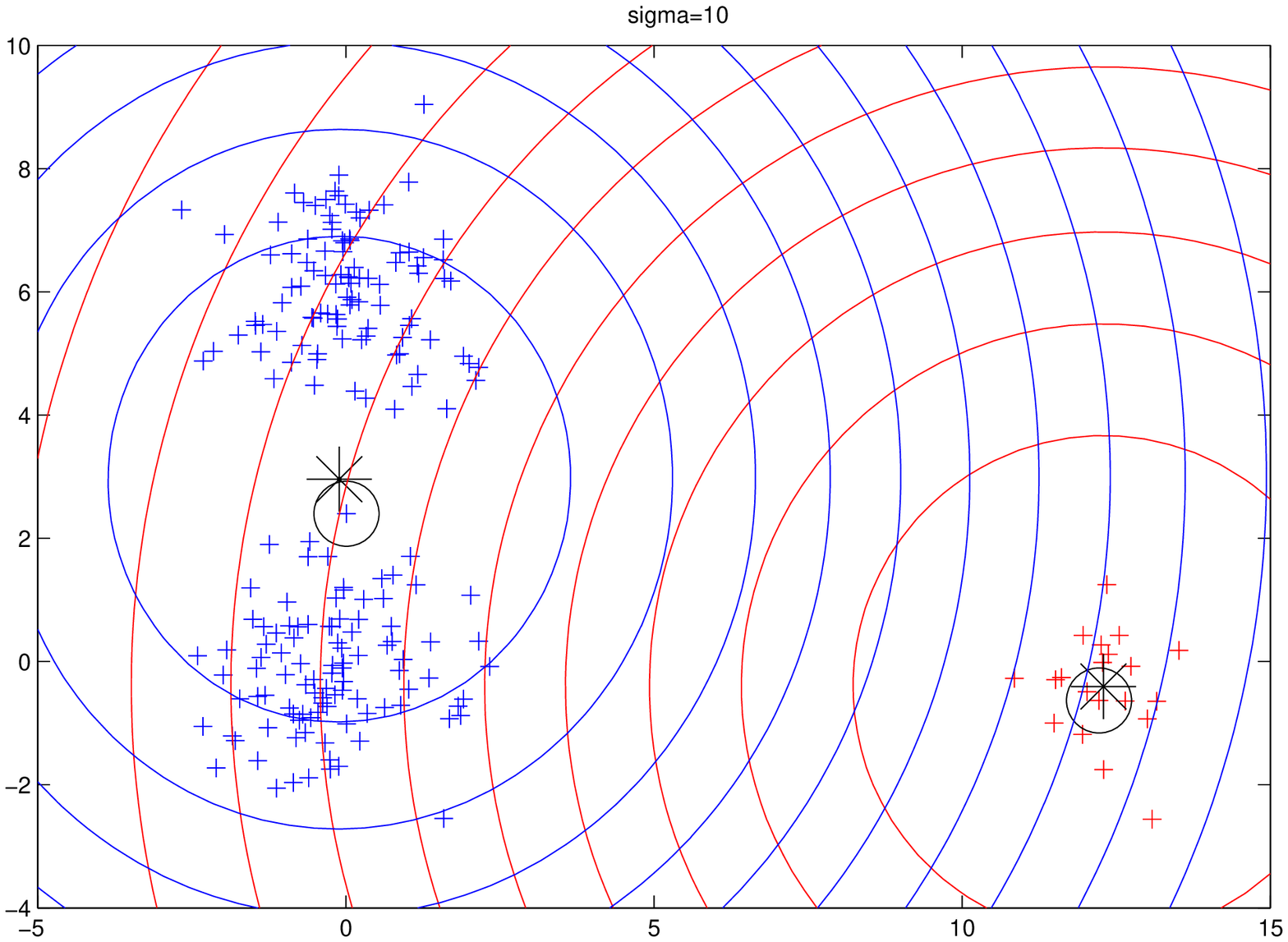} &
      \psfrag{sigma=1}[b][]{$\sigma=1$}
      \includegraphics[width=0.50\linewidth]{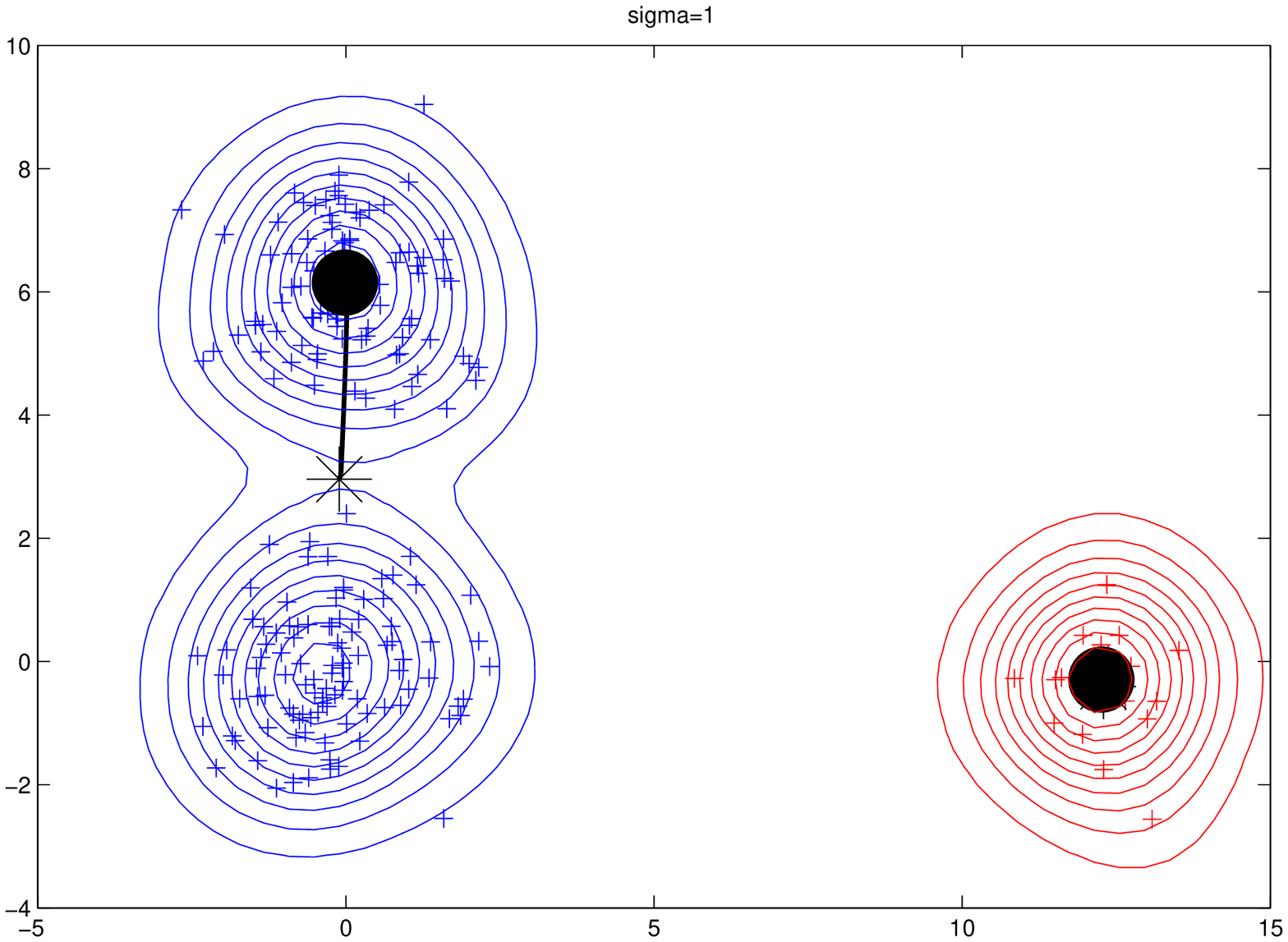}
    \end{tabular}
    \caption{$K$-modes results for two bandwidth values using $K=2$. We show the means $\ast$, their within-cluster nearest neighbor $\circ$, the modes $\bullet$, the paths followed by each mode as $\sigma$ decreases, and the contours of each kde. Each $K$-modes cluster uses a different color.}
    \label{f:3clusters}
  \end{center}
\end{figure}

\begin{figure}[t]
  \begin{center}
    \begin{tabular}{@{}c@{\hspace{0\linewidth}}c@{}}
      \psfrag{sigma=10}[b][]{$\sigma=\infty$ ($K$-means)}
      \includegraphics[width=0.50\linewidth]{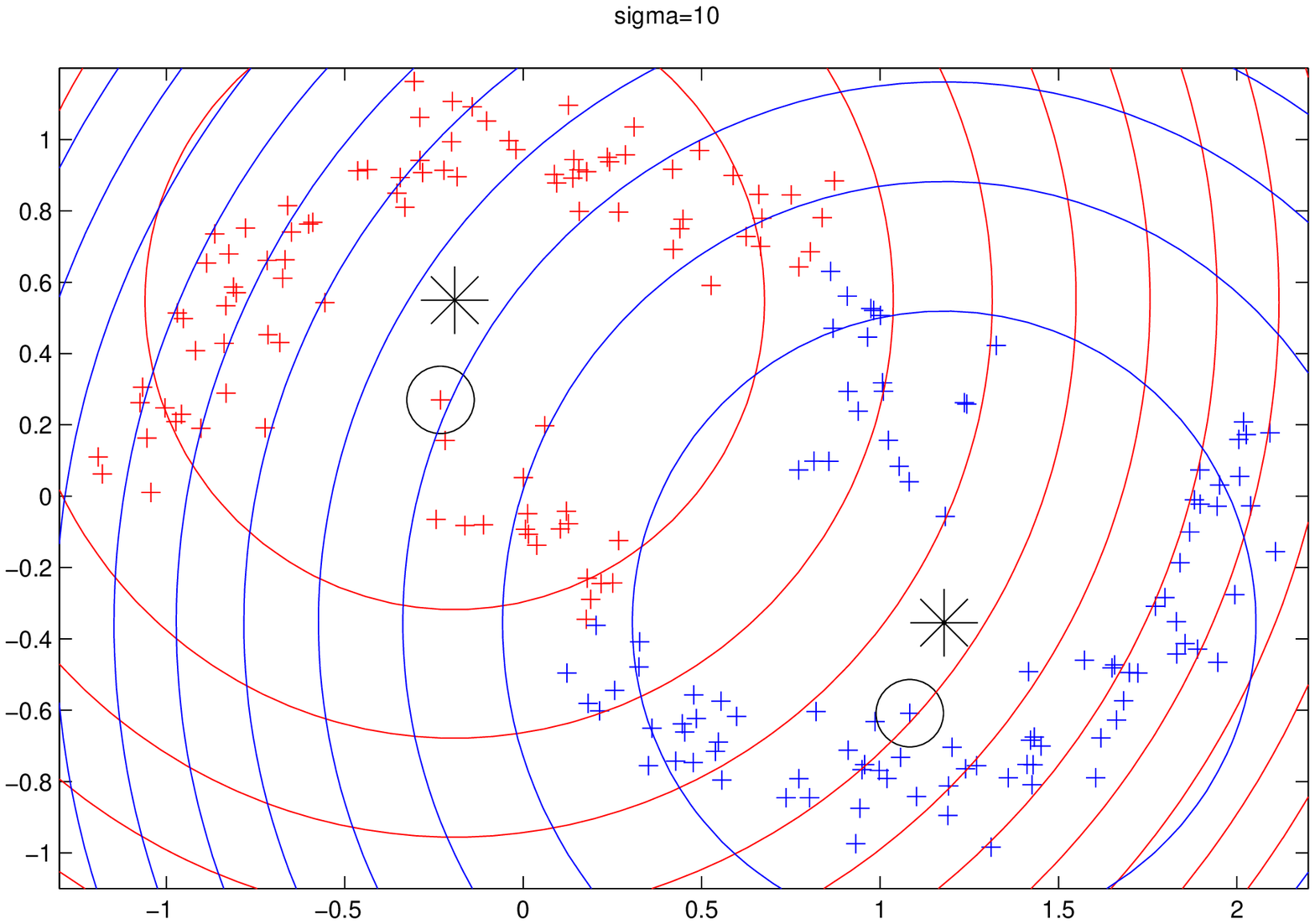} &
      \psfrag{sigma=0.1}[b][]{$\sigma=0.1$}
      \includegraphics[width=0.50\linewidth]{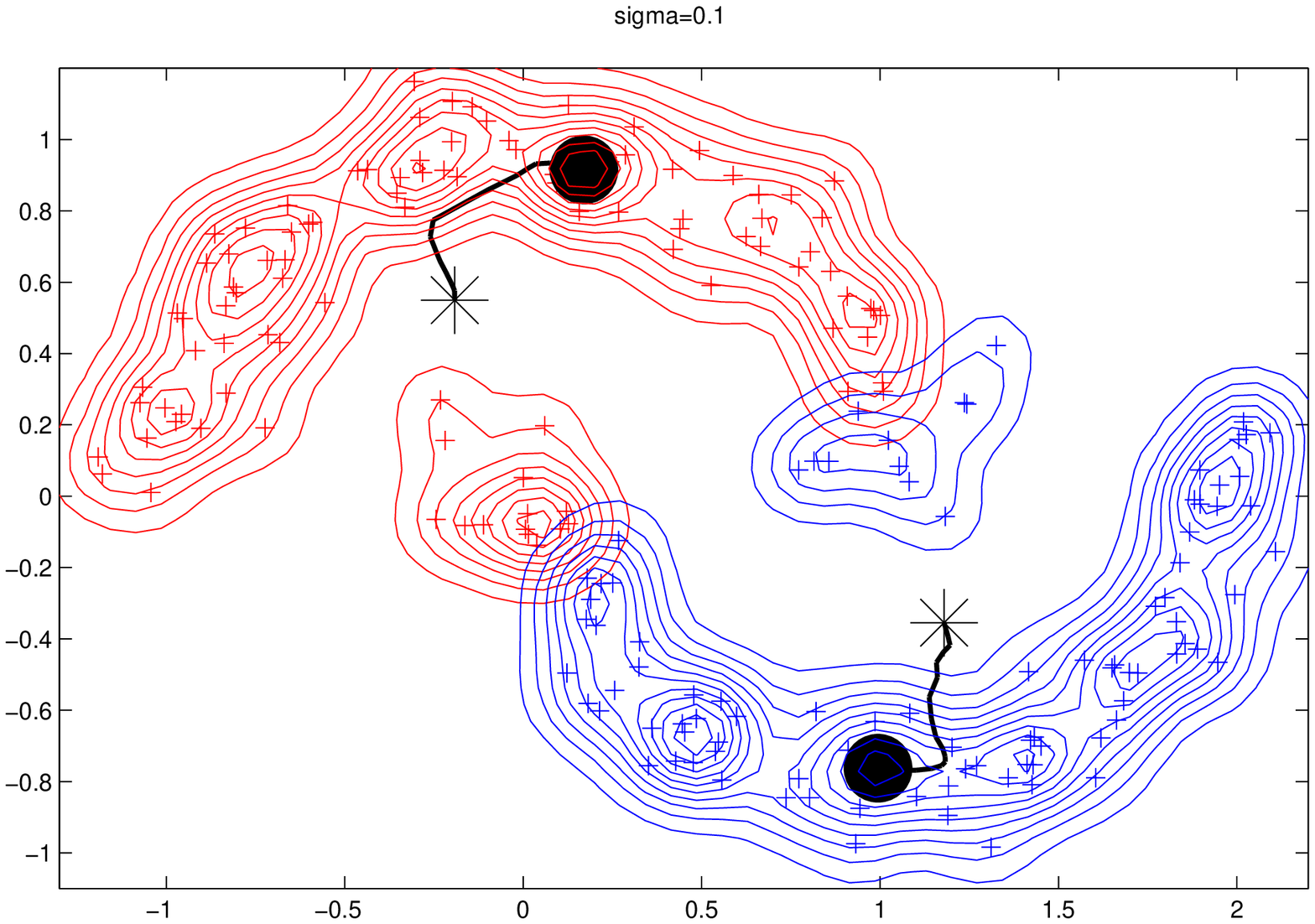}
    \end{tabular}
    \caption{Like fig.~\ref{f:3clusters} but for the two-moons dataset.}
    \label{f:2moons}
  \end{center}
\end{figure}

Figures~\ref{f:3clusters} and~\ref{f:2moons} illustrate the three algorithms in 2D examples. They show the $K$ modes and the kde contours for each cluster, for $\sigma=\infty$ or equivalently $K$-means (left panel) and for an intermediate $\sigma$ (right panel). We run $K$-modes decreasing $\sigma$ geometrically in $20$ steps from $3$ to $1$ in fig.~\ref{f:3clusters} and from $1$ to $0.1$ in fig.~\ref{f:2moons}.

In fig.~\ref{f:3clusters}, which has 3 Gaussian clusters, we purposefully set $K=2$ (both $K$-means and $K$-modes work well with $K=3$). This makes $K$-means put one of the centroids in a low-density area, where no input patterns are found. $K$-modes moves the centroid inside a cluster in a maximum-density area, where many input patterns lie, and is then more representative.

In fig.~\ref{f:2moons}, the ``two-moons'' dataset has two nonconvex, interleaved clusters and we set $K=2$. The ``moons'' cannot be perfectly separated by either $K$-means or $K$-modes, since both define Voronoi tessellations. However, $K$-modes does improve the clusters over those of $K$-means, and as before it moves the centroids from a region where no patterns are found to a more typical location within each cluster. Note how, although the bandwidth used ($\sigma = 0.1$) yields a very good kde for each cluster and would also yield a very good kde for the whole dataset, it results in multiple modes for each ``moon'', which means that GMS would return around 13 clusters. In this dataset, no value of $\sigma$ results in two modes that separate the moons.

One might argue that, if a $K$-means centroid is not a valid pattern, one could simply replace it with the data point from its cluster that is closest to it. While this sometimes works, as would be the case in the rotated-digit-1 of fig.~\ref{f:rotated1}, it often does not: the same-cluster nearest neighbor could be a point on the cluster boundary, therefore atypical (fig.~\ref{f:3clusters}) or even a point in the wrong cluster (fig.~\ref{f:2moons}). $K$-modes will find points interior to the clusters, with higher density and thus more typical.

\subsubsection*{Degree Distribution of a Graph}

\begin{figure}[p]
  \begin{center}
    \begin{tabular}{@{\hspace{0\linewidth}}c@{\hspace{0\linewidth}}c@{\hspace{0\linewidth}}c@{}}
      histogram & per-cluster kde & whole-data kde \\
      \psfrag{sigma=200}[b][]{}
      \psfrag{histogram}[][t]{$\sigma\rightarrow\infty$ ($K$-means)}
      \psfrag{degree}{}
      \includegraphics[width=0.32\linewidth]{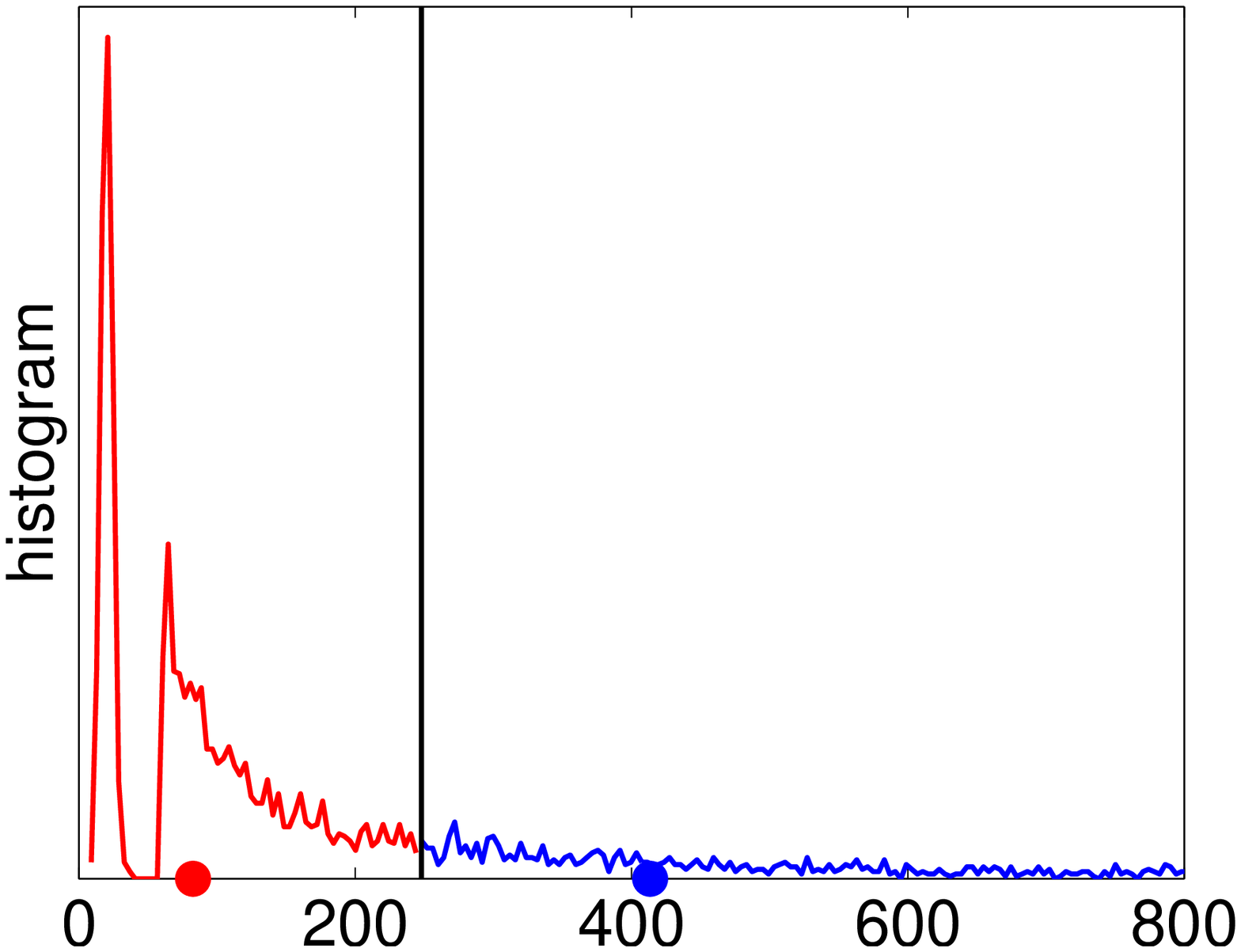} &
      \psfrag{kde}{}
      \psfrag{degree}{}
      \includegraphics[width=0.30\linewidth]{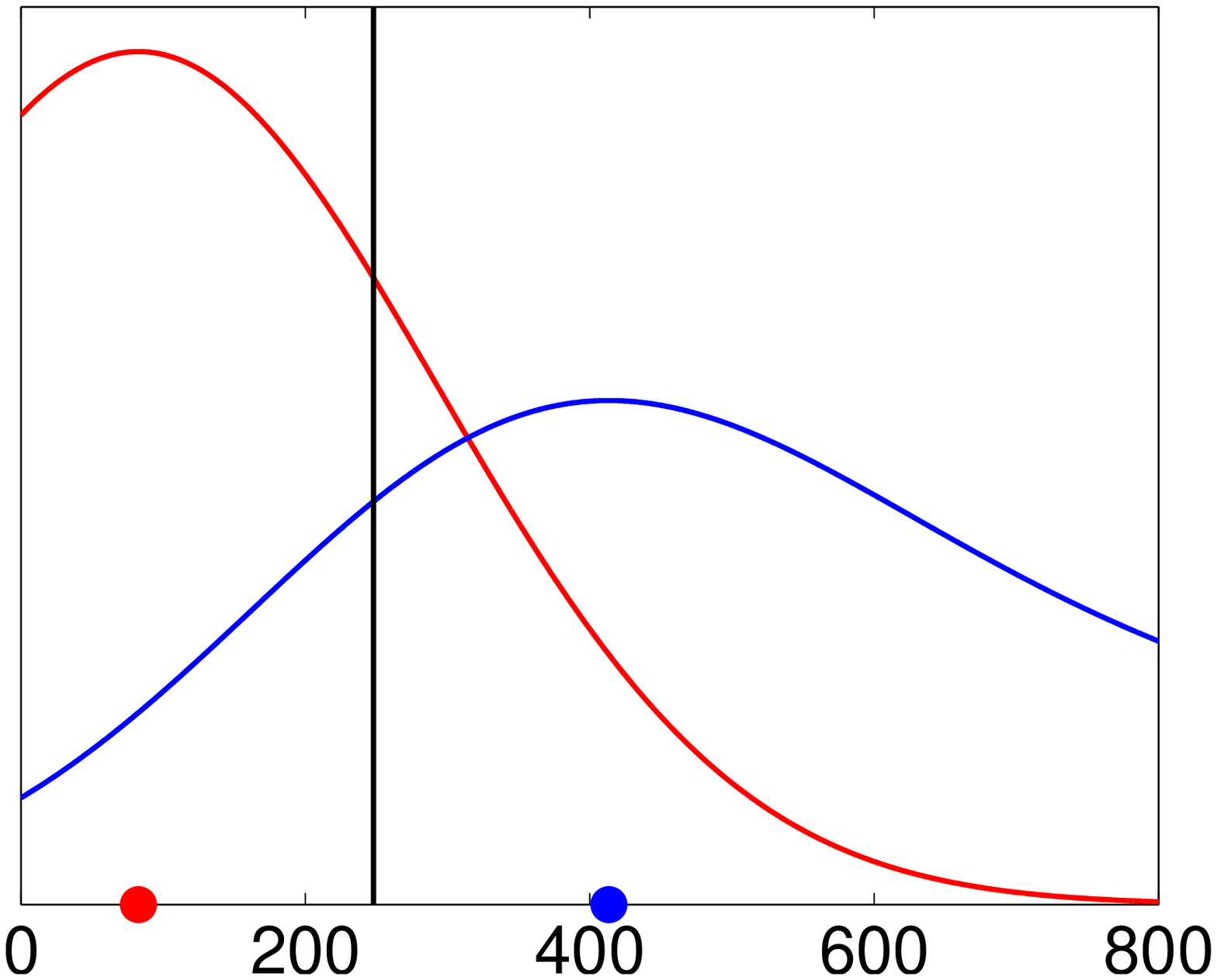} &
      \psfrag{kde}{}
      \psfrag{degree}{}
      \includegraphics[width=0.30\linewidth]{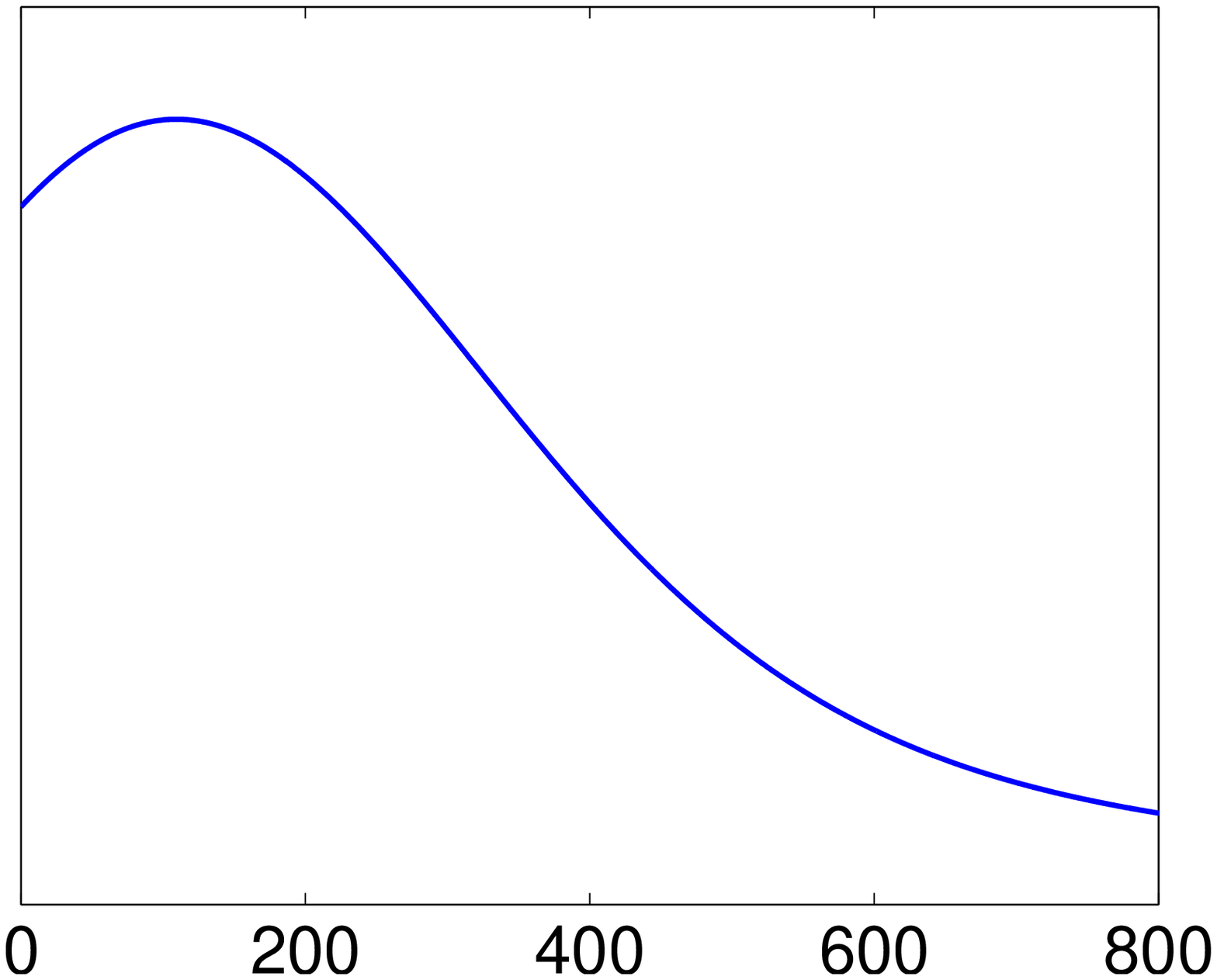} \\
      \psfrag{sigma=29.8551}[b][]{}
      \psfrag{histogram}[][t]{$\sigma=29.86$}
      \psfrag{degree}{}
      \includegraphics[width=0.32\linewidth]{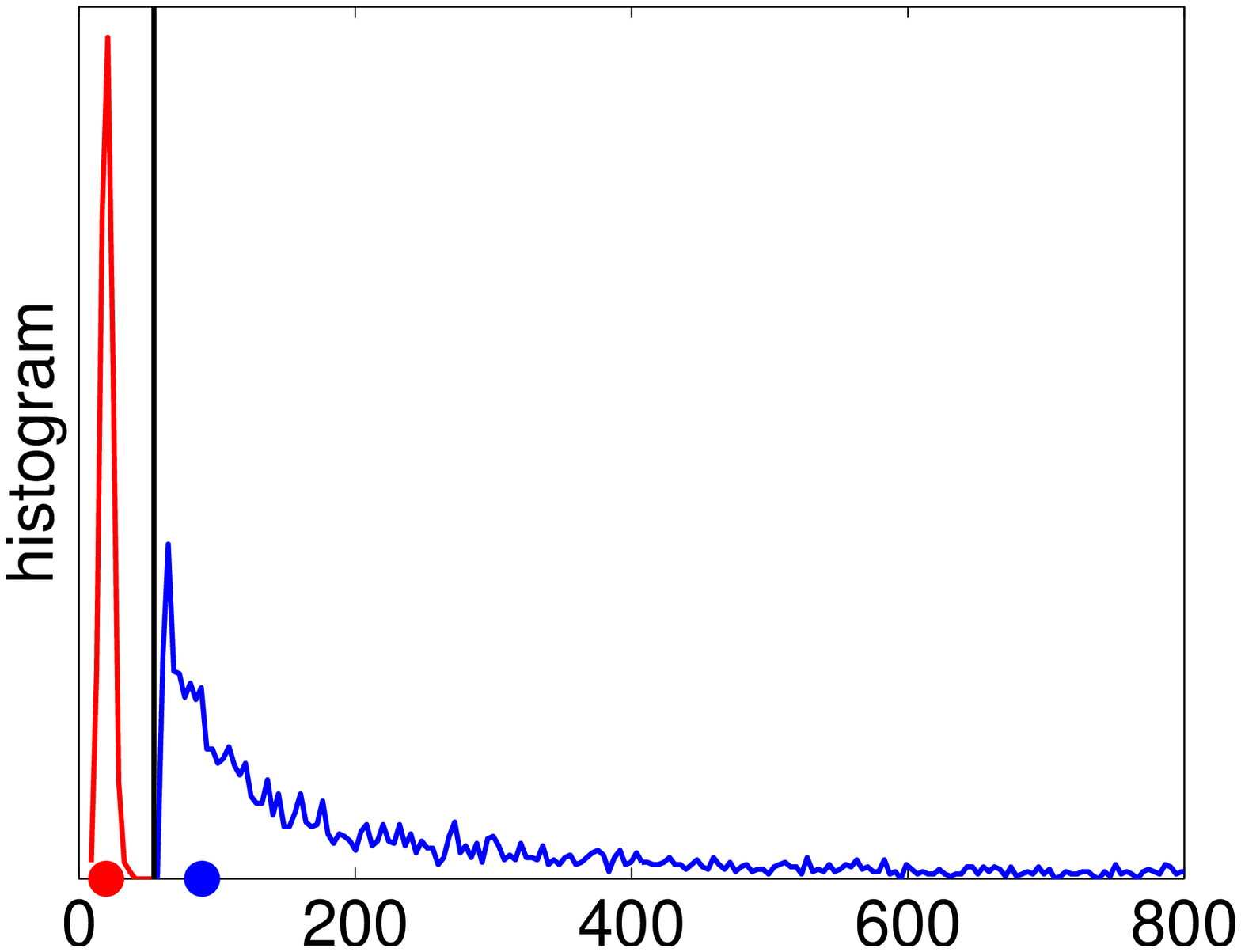} &
      \psfrag{degree}{}
      \includegraphics[width=0.30\linewidth]{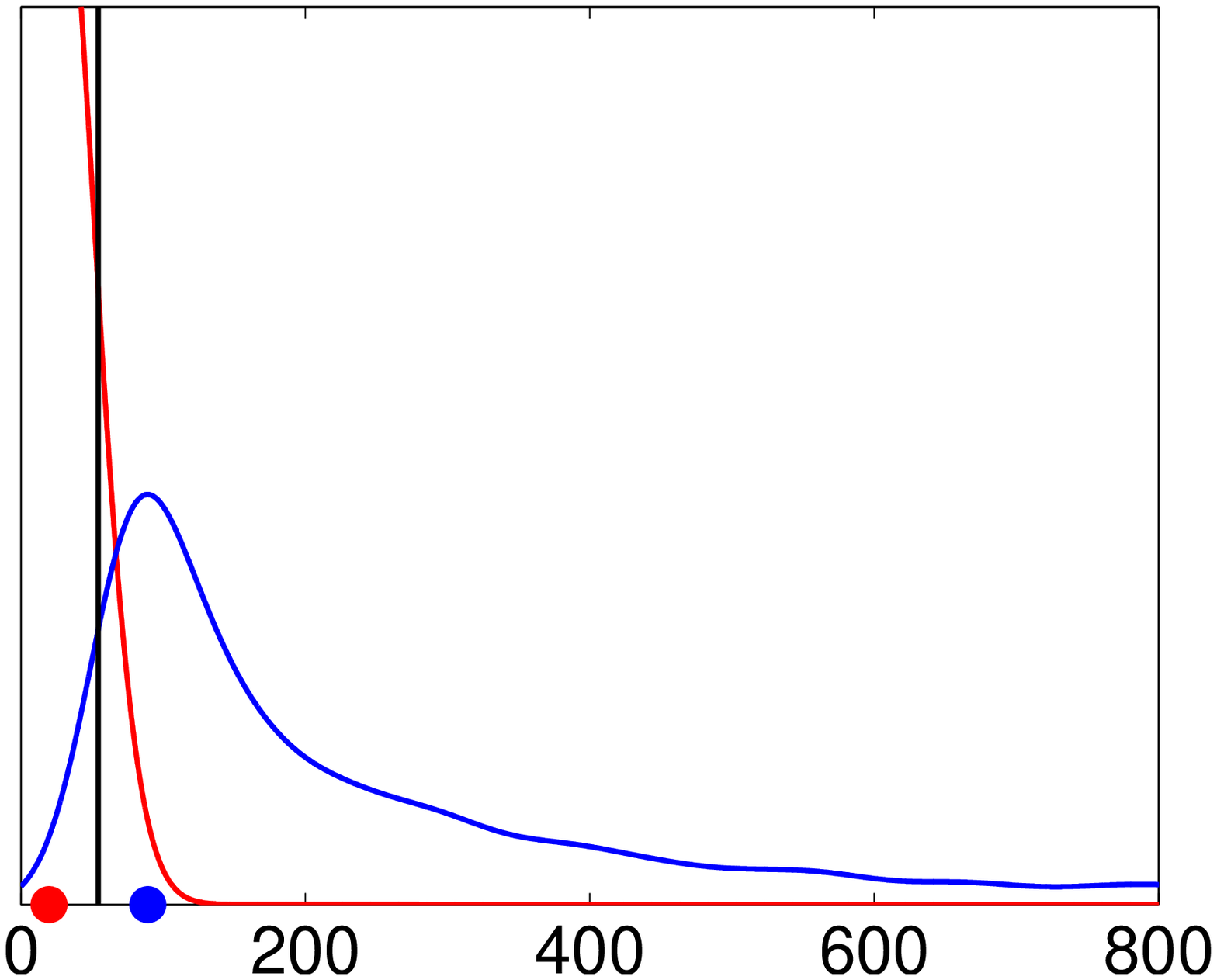} &
      \psfrag{degree}{}
      \includegraphics[width=0.30\linewidth]{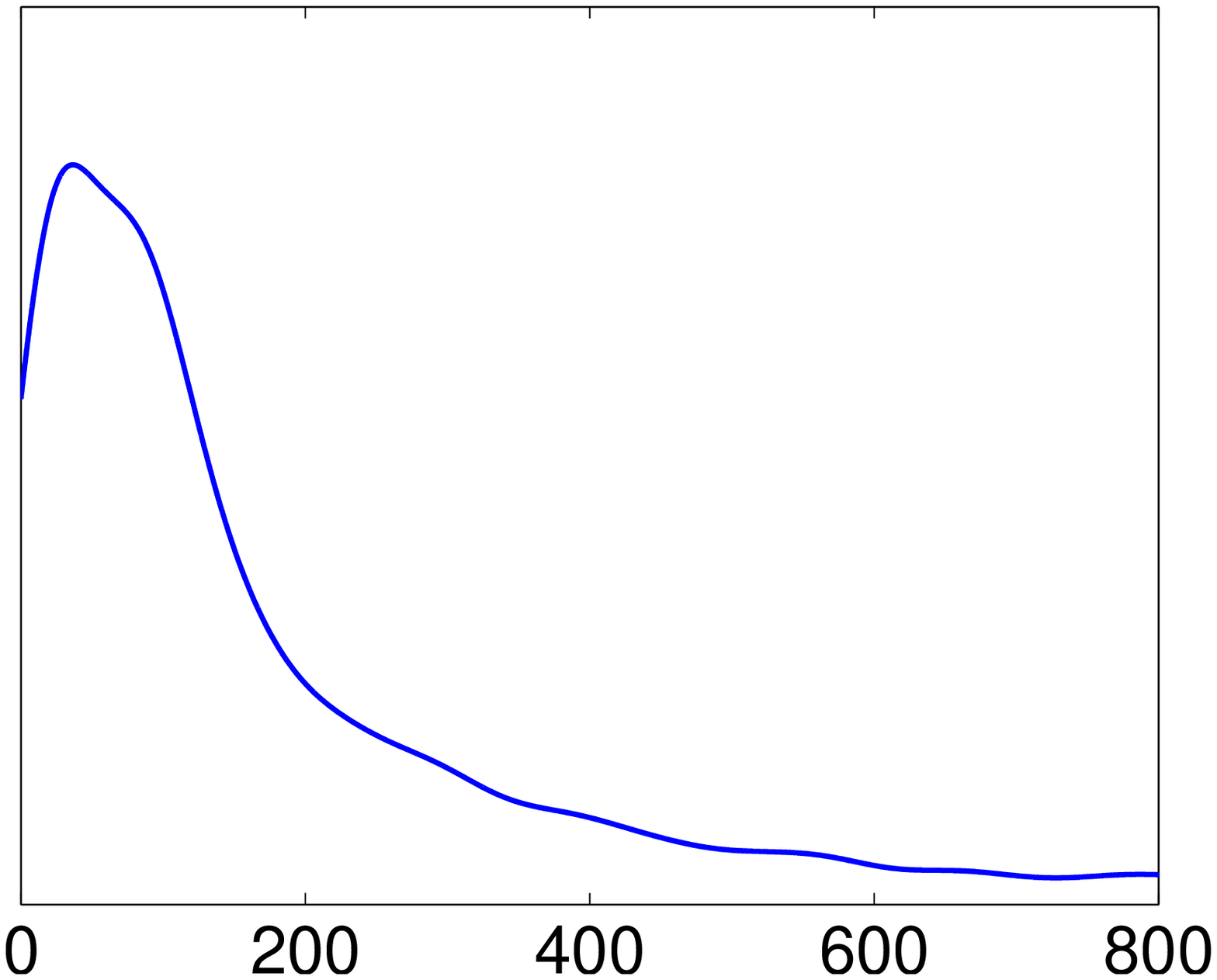} \\
      \psfrag{sigma=15.1361}[b][]{}
      \psfrag{histogram}[][t]{$\sigma=15.14$}
      \psfrag{degree}{}
      \includegraphics[width=0.32\linewidth]{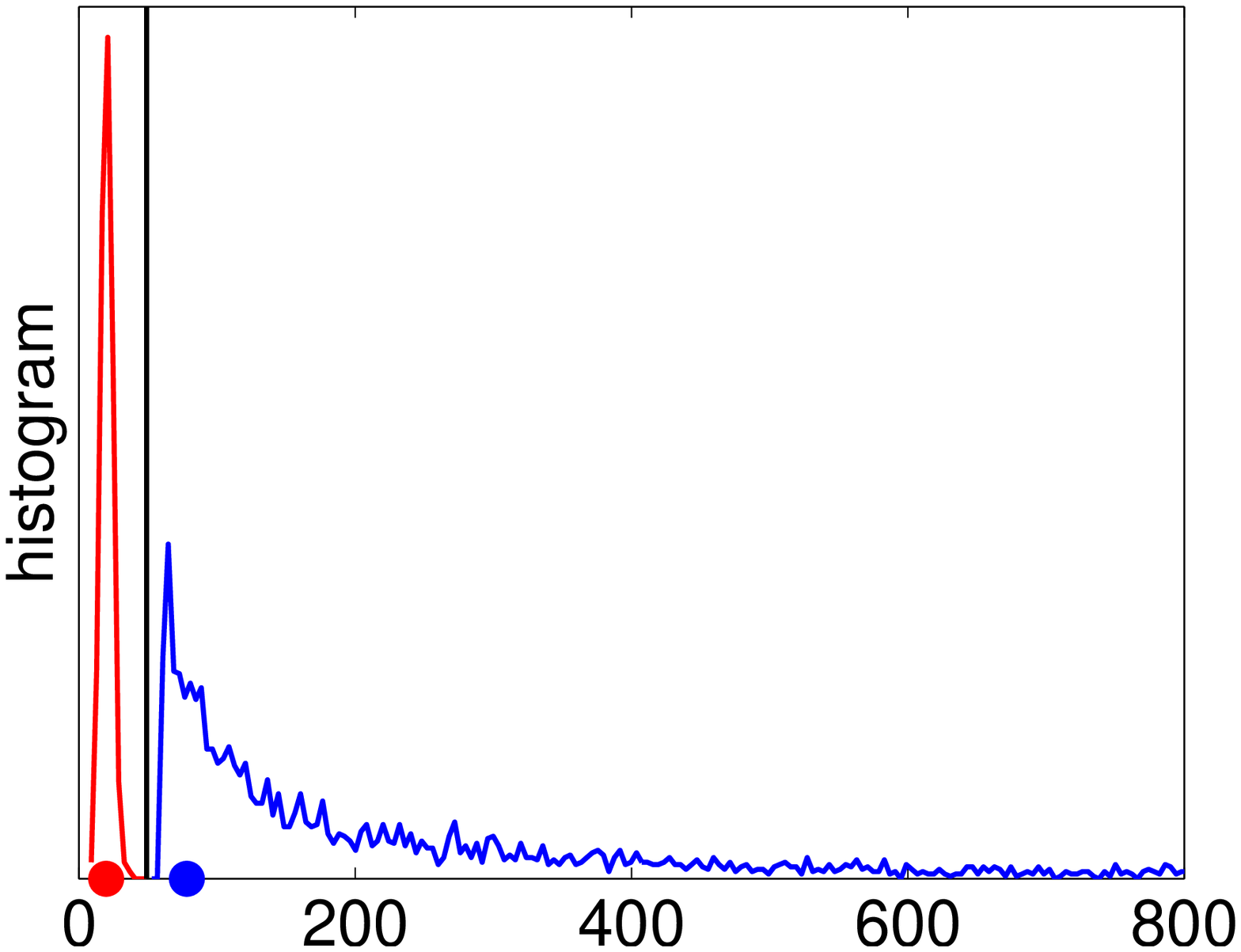} &
      \psfrag{degree}{}
      \includegraphics[width=0.30\linewidth]{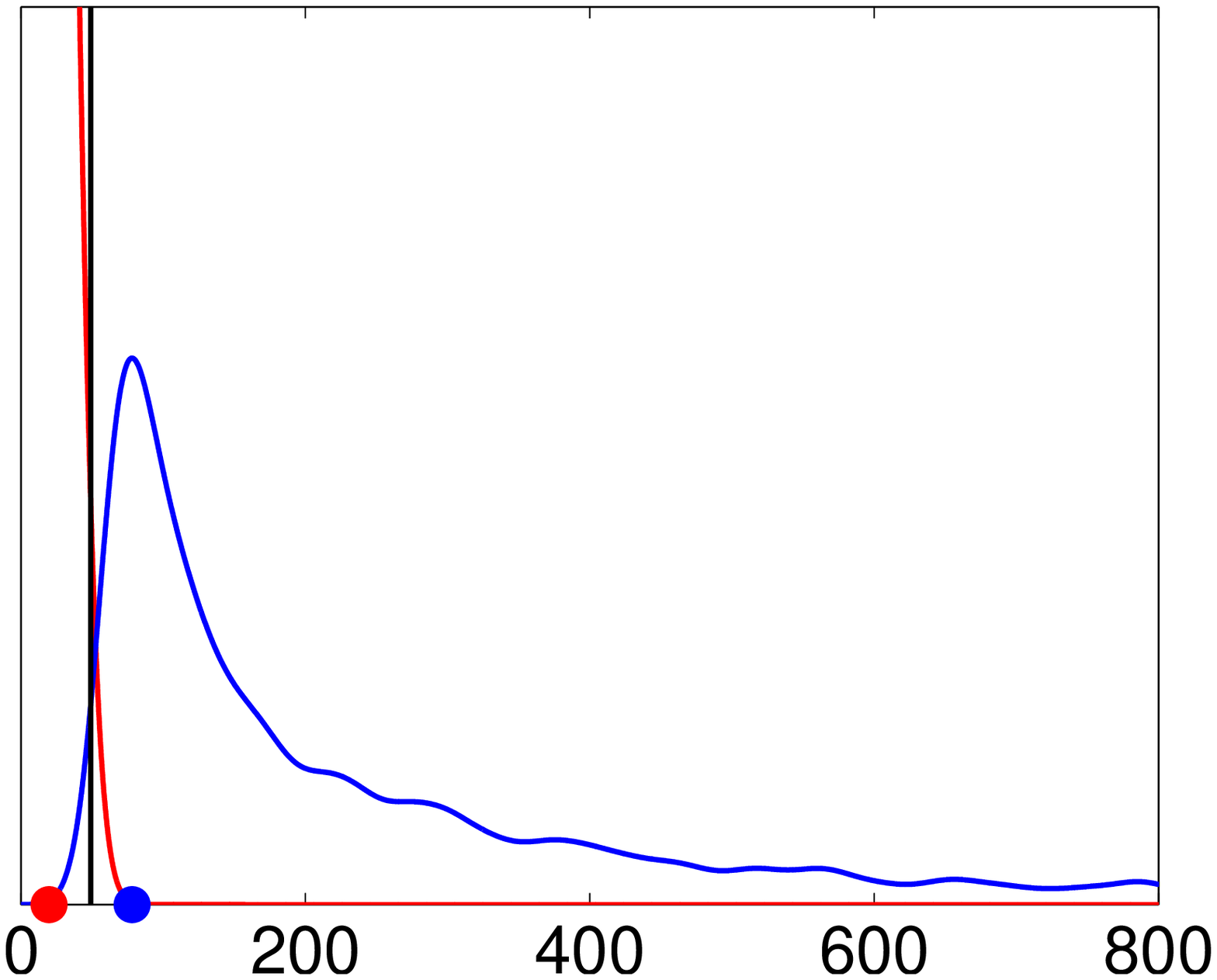} &
      \psfrag{degree}{}
      \includegraphics[width=0.30\linewidth]{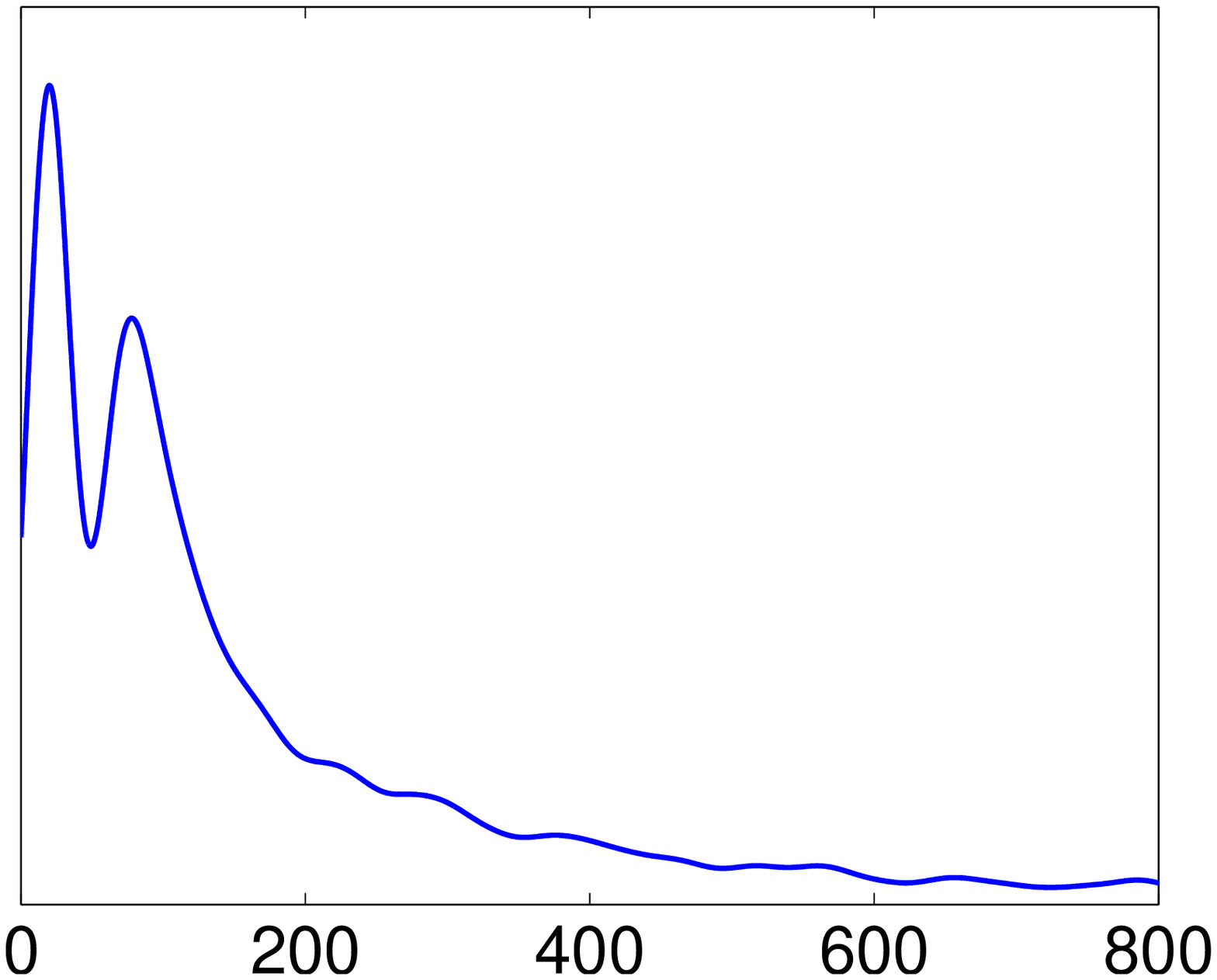} \\
      \psfrag{sigma=3.8905}[b][]{}
      \psfrag{histogram}[][t]{$\sigma=3.89$}
      \psfrag{degree}{}
      \includegraphics[width=0.32\linewidth]{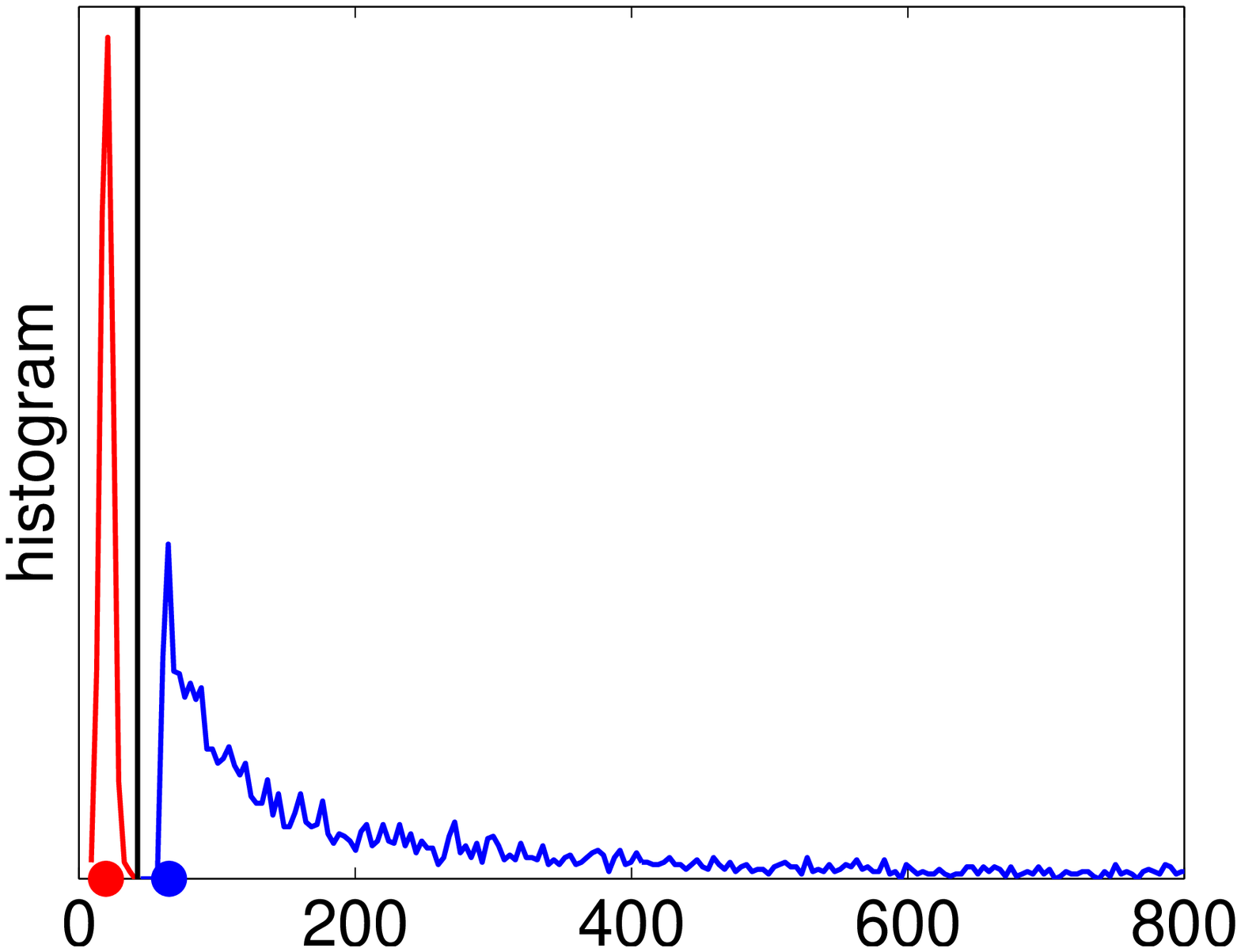} &
      \psfrag{degree}{}
      \includegraphics[width=0.30\linewidth]{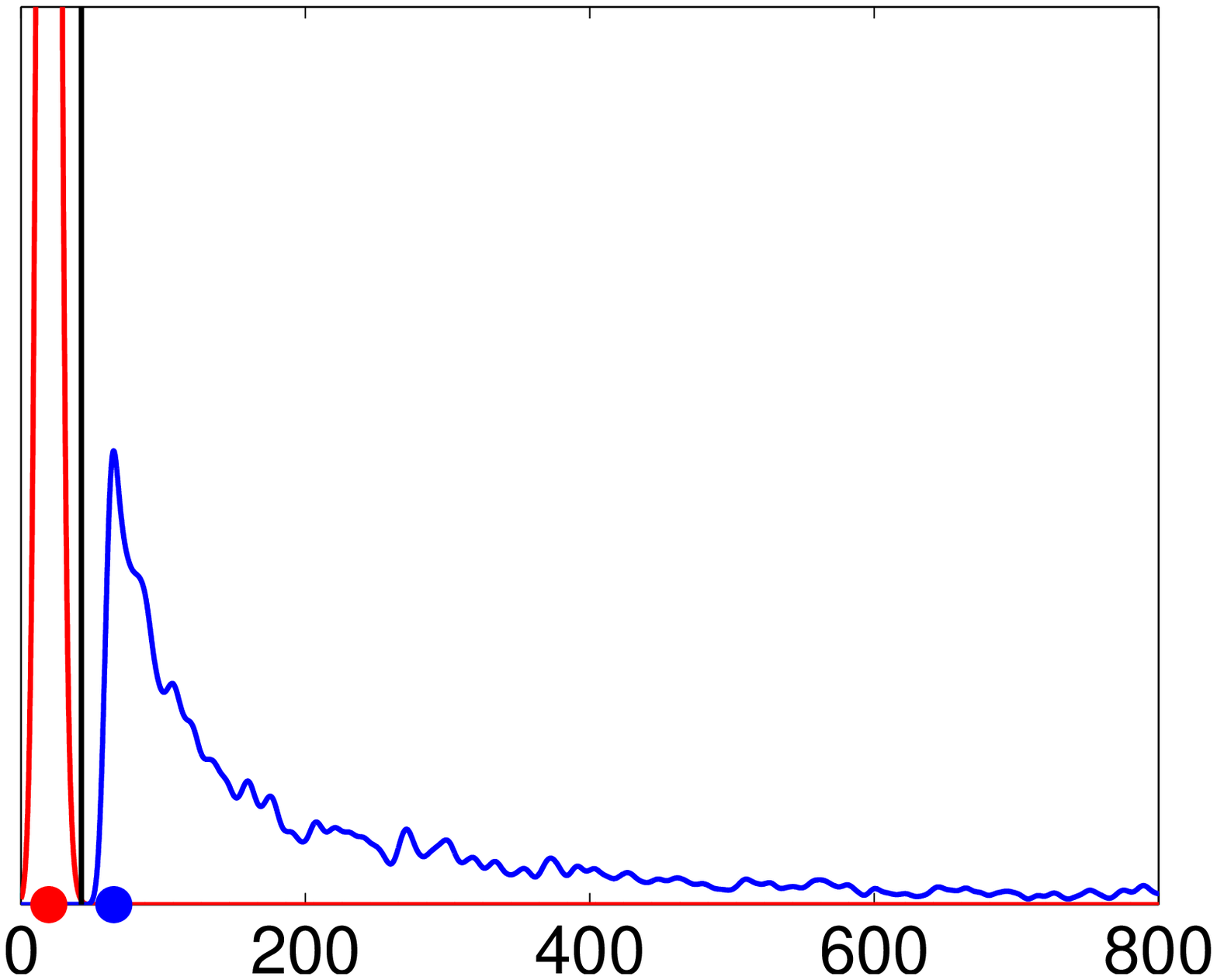} &
      \psfrag{degree}{}
      \includegraphics[width=0.30\linewidth]{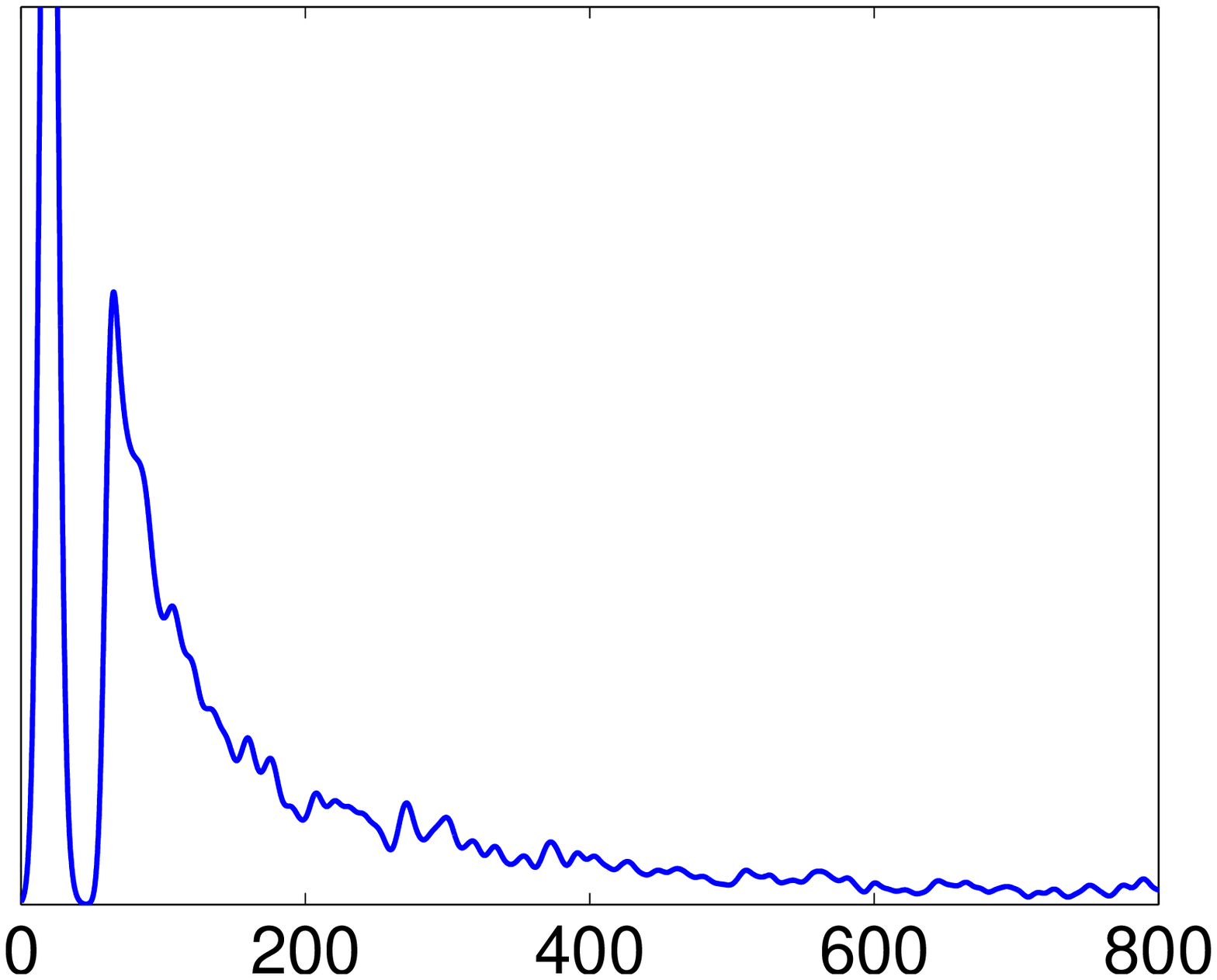} \\
      \psfrag{sigma=1}[b][]{}
      \psfrag{histogram}[][t]{$\sigma=1$}
      \psfrag{degree}[t][]{degree}
      \includegraphics[width=0.32\linewidth]{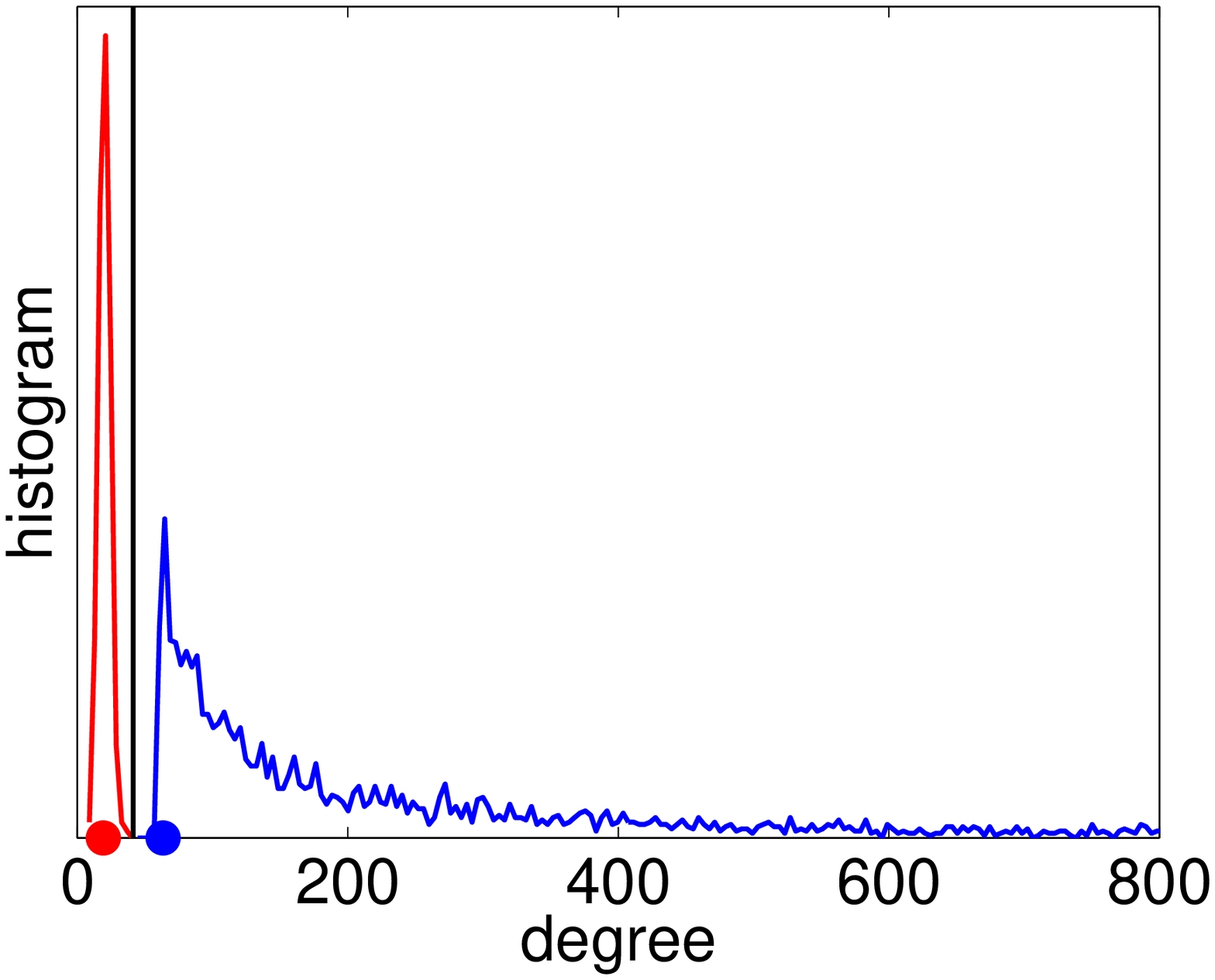} &
      \psfrag{sigma=1}[b][]{}
      \psfrag{degree}[t][]{degree}
      \includegraphics[width=0.30\linewidth]{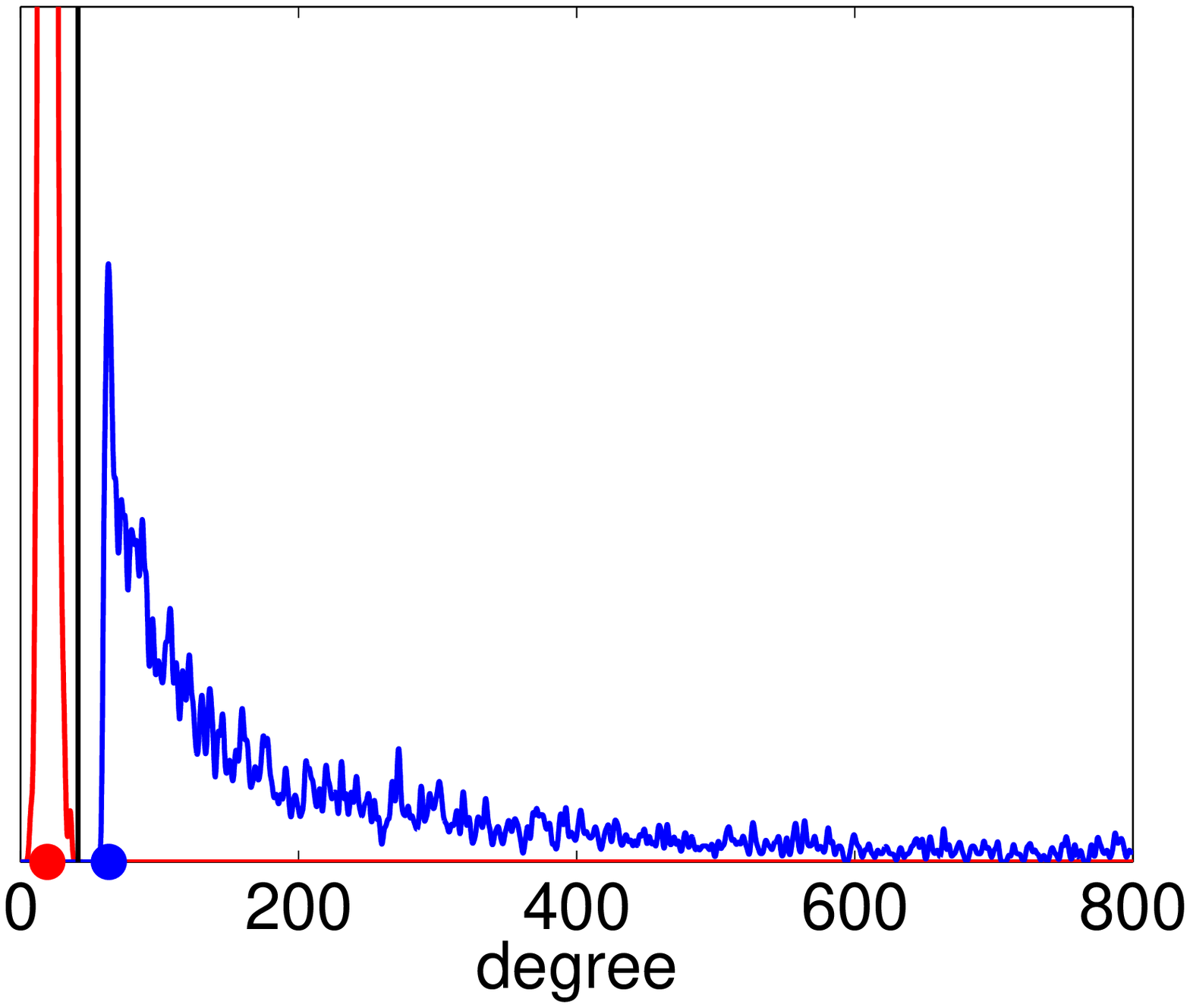} &
      \psfrag{degree}[t][]{degree}
      \includegraphics[width=0.30\linewidth]{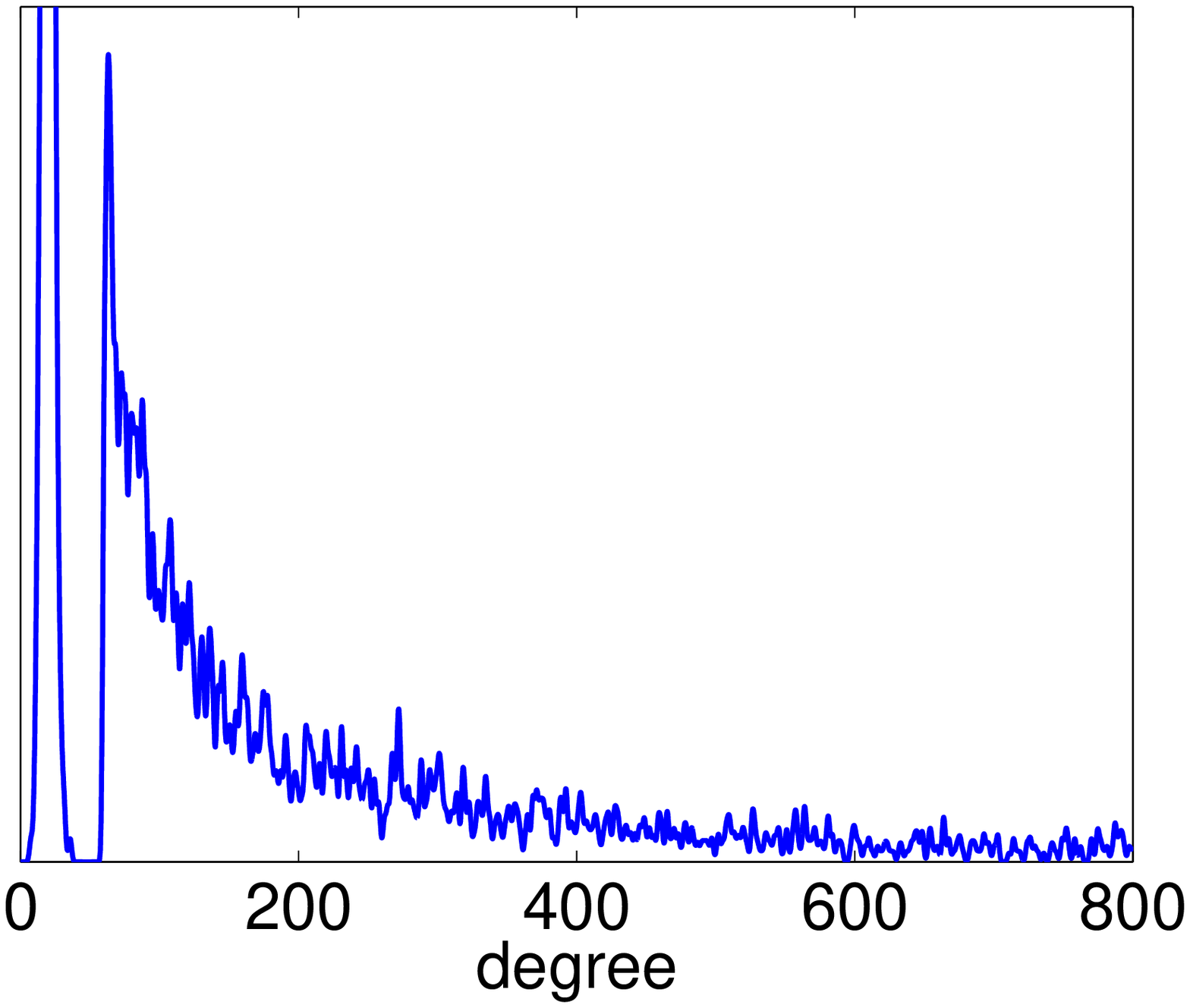}
    \end{tabular}
    \caption{Degree distribution of a graph. \emph{Left column}: a histogram of the distribution, colored according to the $K$-modes clustering for $\sigma=\infty$ ($K$-means) to $\sigma=1$; the black vertical bar indicates the cluster boundary. \emph{Middle column}: the kde for each cluster with $K$-modes. \emph{Right column}: the kde for the whole dataset with GMS. The X axis is truncated to a degree of 800, so many outlying modes to the right are not shown.}
    \label{f:degrees}
  \end{center}
\end{figure}

We construct an undirected graph similar to many real-world graphs and estimate the distribution of the degree of each vertex \cite{Newman10a}. To construct the graph, we generated a random (Erd{\H{o}}s-R{\'e}nyi) graph (with $1\,000$ vertices and $9\,918$ edges), which has a Gaussian degree distribution, and a graph with a power-law (long-tailed) distribution (with $3\,000$ vertices and $506\,489$ edges), and then took the union of both graphs and added a few edges at random connecting the two subgraphs. The result is a connected graph with two types of vertices, reminiscent of real-world networks such as the graph of web pages and their links in the Internet. Thus, our dataset has $N=4\,000$ points in 1D (the degree of each vertex). As shown in fig.~\ref{f:degrees}, the degree distribution is a mixture of two distributions that are well-separated but have a very different character: a Gaussian and a skewed, power-law distribution. The latter results in a few vertices having a very large degree (e.g.\ Internet hubs), which practically appear as outliers to the far right (outside the plots).

We set $K=2$. $K$-means obtains a wrong clustering. One centroid is far to the right, in a low-density (thus unrepresentative) region, and determines a cluster containing the tail of the power-law distribution; this is caused by the outliers. The other centroid is on the head of the power-law distribution and determines a cluster containing the Gaussian and the head of the power-law distribution.

We run $K$-modes decreasing $\sigma$ from $200$ to $1$ geometrically in $40$ steps. $K$-modes shifts the centroids to the two principal modes of the distributions and achieves a perfect clustering. Note that the kde for the power-law cluster has many modes, but $K$-modes correctly converges to the principal one.

GMS cannot separate the two distributions for any value of $\sigma$. Setting $\sigma$ small enough that the kde has the two principal modes implies it also has many small modes in the tail because of the outliers (partly visible in the second panel). This is a well-known problem with kernel density estimation.

\subsubsection*{Handwritten Digit Images}

We selected 100 random images ($16\times 16$ grayscale) from the USPS dataset for each digit 0--9. This gives a dataset of $N=1\,000$ points in $[0,1]^{256}$. We ran $K$-means and $K$-modes with $K=10$, decreasing $\sigma$ from $10$ to $1$ geometrically in $100$ steps.

\begin{figure}[p]
  \begin{center}
    \begin{tabular}{@{}c@{}}
      \psfrag{sigma=Inf}[][l]{$K$-modes $\sigma=\infty$ ($K$-means)}
      \includegraphics[width=\linewidth]{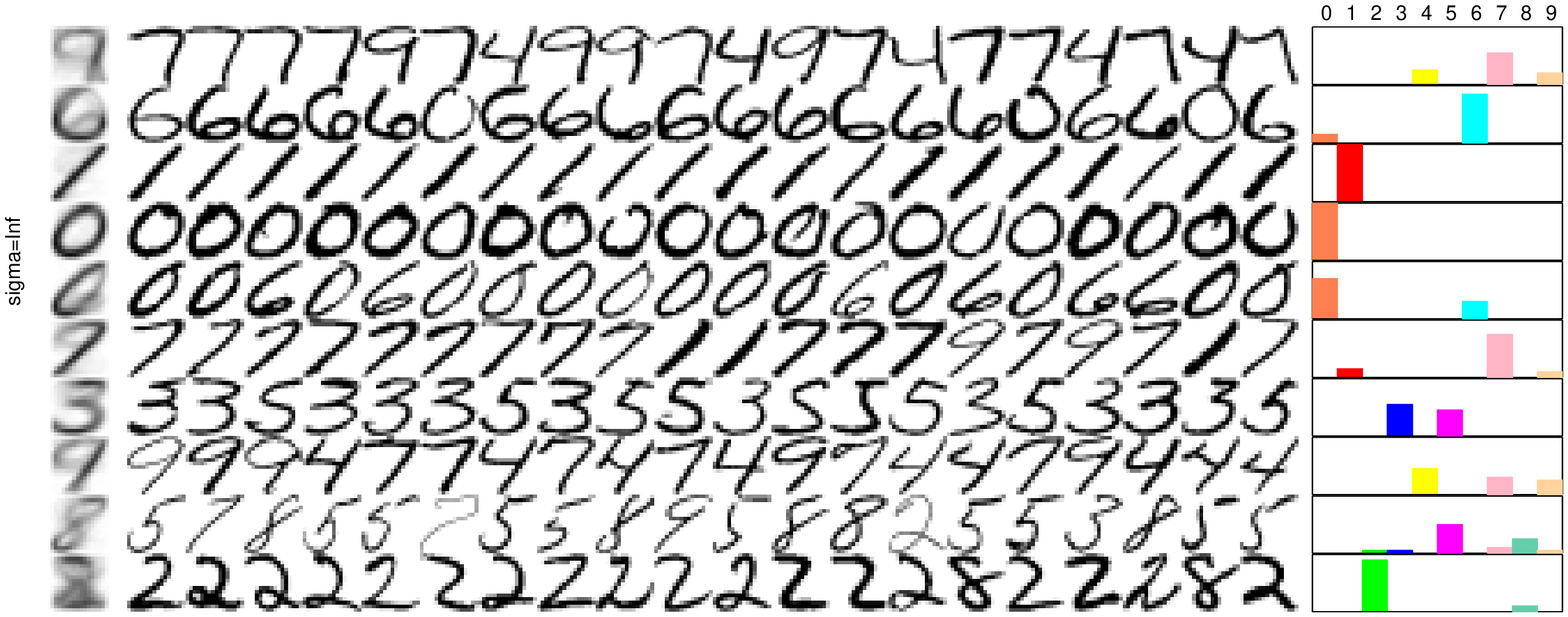} \\[2ex]
      \psfrag{sigma=1}[][l]{$K$-modes $\sigma=1$}
      \includegraphics[width=\linewidth]{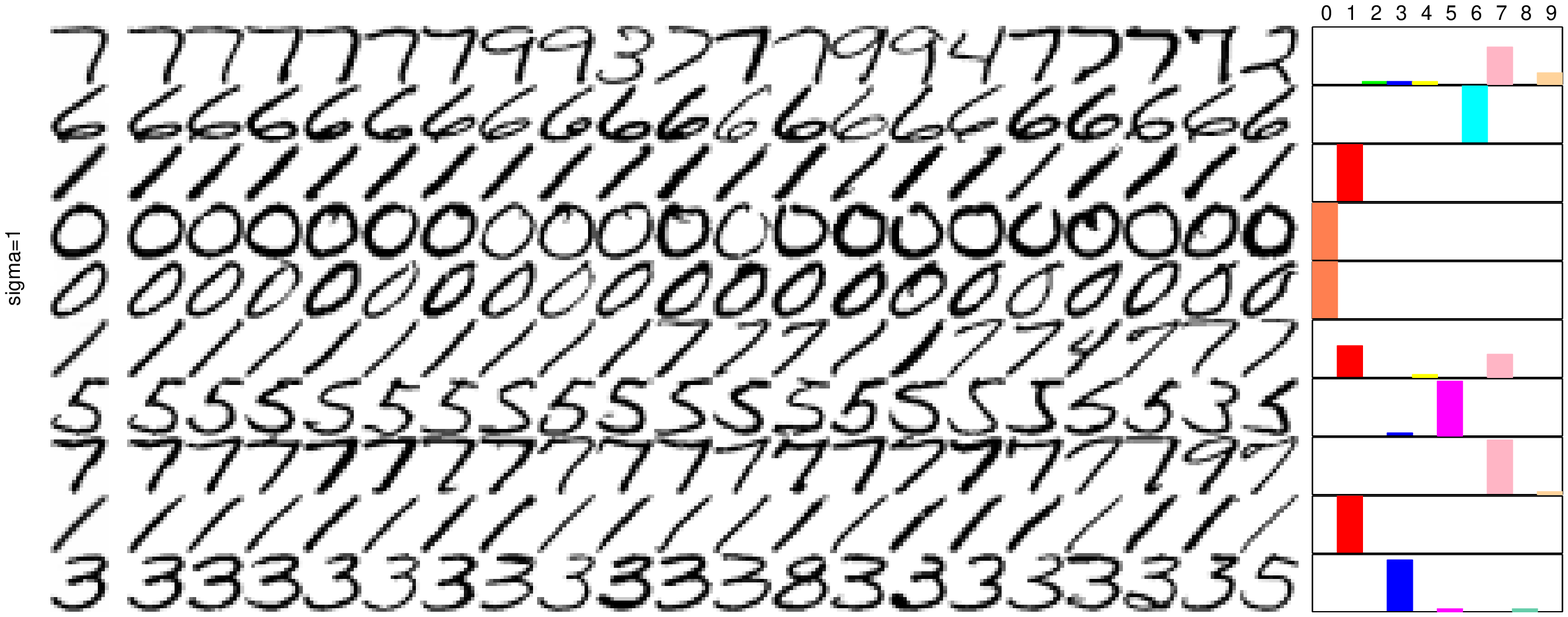} \\[2ex]
      \psfrag{sigma=1.8369}[][l]{GMS $\sigma=1.8369$}
      \includegraphics[width=\linewidth]{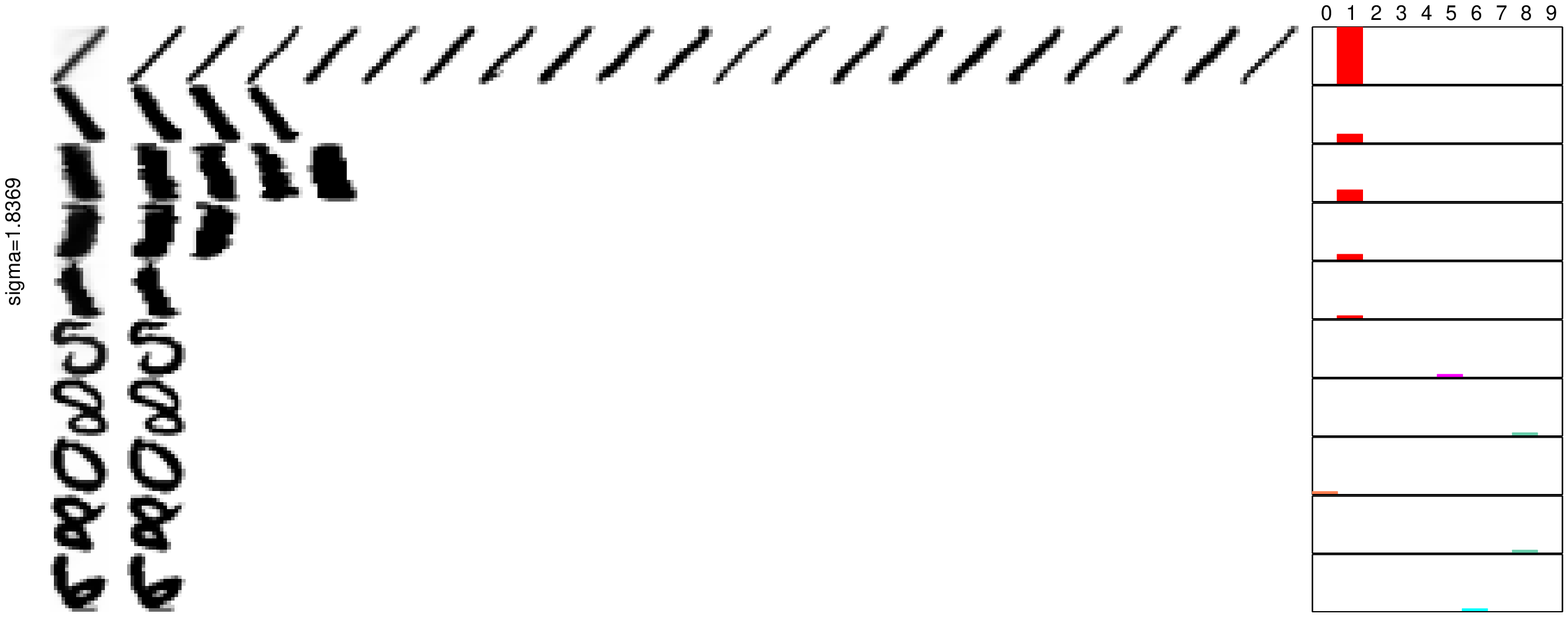}
    \end{tabular}
    \caption{Clustering results on USPS data with $K$-modes with $K=10$ for $\sigma=\infty$ (i.e., $K$-means, top panel) and $\sigma=1$ (middle panel), and for GMS with $\sigma=1.8369$ (bottom panel), which achieves $K=10$ modes. In each panel, each row corresponds to a cluster $k=1,\dots,K=10$. The leftmost image shows the centroid $\c_k$ and the right 20 images are the 20 nearest neighbors to it within cluster $k$. The right panel shows the histogram of class labels (color-coded) for the neighbors.}
    \label{f:usps}
  \end{center}
\end{figure}

Fig.~\ref{f:usps} shows that most of the centroids for $K$-means are blurry images consisting of an average of digits of different identity and style (slant, thickness, etc.), as seen from the 20 nearest-neighbor images of each centroid (within its cluster). Such centroids are hard to interpret and are not valid digit images. This also shows how the nearest neighbor to the centroid may be an unusual or noisy input pattern that is not representative of anything except itself.

$K$-modes unblurs the centroids as $\sigma$ decreases. The class histograms for the 20 nearest-neighbors show how the purity of each cluster improves: for $K$-means most histograms are widely distributed, while $K$-modes concentrates the mass into mostly a single bin. This means $K$-modes moves the centroids onto typical regions that both look like valid digits, and are representative of their neighborhood. This can be seen not just from the class labels, but also from the style of the digits, which becomes more homogeneous under $K$-modes (e.g.\ see cluster $\c_2$, containing digit-6 images, or $\c_4$ and $\c_5$, containing digit-0 images of different style).

Stopping $K$-modes at an intermediate $\sigma$ (preventing it from becoming too small) achieves just the right amount of smoothing. It allows the centroids to look like valid digit images, but at the same time to average out noise, unusual strokes or other idiosyncrasies of the dataset images (while not averaging digits of different identities or different styles, as $K$-means does). This yields centroids that are more representative even than individual images of the dataset. In this sense, $K$-modes achieves a form of intelligent denoising similar to that of manifold denoising algorithms \cite{WangCarreir10a}.

Note that, for $K$-modes, centroids $\c_6$ and $\c_9$ look very similar, which suggests one of them is redundant (while none of the $K$-means centroids looked very similar to each other). Indeed, removing $\c_6$ and rerunning $K$-modes with $K=9$ simply reassigns nearly all data points in the cluster of $\c_6$ to that of $\c_9$ and the centroid itself barely changes. This is likely not a casuality. If we have a single Gaussian cluster but use $K>1$, it will be split into sectors like a pie, but in $K$-means the centroids will be apart from each other, while in $K$-modes they will all end up near the Gaussian center, where the mode of each kde will lie. This suggests that redundancy may be easier to detect in $K$-modes than in $K$-means.

GMS with $\sigma=1.8369$ gives exactly $10$ modes, however of these one is a slanted-digit-1 cluster like $\c_9$ in $K$-modes and contains $98.5$\% of the training set points, and the remaining 9 modes are associated with clusters containing between 1 and 4 points only, and their centroids look like digits with unusual shapes, i.e., outliers. As noted before, GMS is sensitive to outliers, which create modes at nearly all scales. This is particularly so with high-dimensional data, where data is always sparse, or with data lying on a low-dimensional manifold (both of which occur here). In this case, the kde changes from a single mode for large $\sigma$ to a multiplicity of modes over a very narrow interval of $\sigma$.

\subsubsection*{High-dimensional Datasets with Ground-Truth Labels}

Finally, we report clustering statistics in datasets with known pattern class labels (which the algorithms did not use): (1) COIL--20, which contains $32\times 32$ grayscale images of 20 objects viewed from varying angles. (2) MNIST, which contains $28 \times 28$ grayscale handwritten digit images (we randomly sample 200 of each digit). And (3) the NIST Topic Detection and Tracking (TDT2) corpus, which contains on-topic documents of different semantic categories (we removed documents appearing in more than one category and kept only the largest 30 categories).

For each algorithm, we compare its clustering with the ground-truth one using two commonly used criteria for evaluating clustering results: the Adjusted Rand Index and the Normalized Mutual Information \cite{Mannin_08a}. The results appear in table~\ref{t:hd-datasets}. For $K$-means, we show the best result of 20 random initializations. $K$-modes was initialized from this $K$-means result and run by homotopy to a target bandwidth $\sigma^*$. We show two results for $K$-modes, each with a different bandwidth value. The one in parentheses corresponds to a target $\sigma^*$ estimated as the average distance of each point to its $10$th nearest neighbor (a commonly used bandwidth estimation rule). The other one (not in parentheses) corresponds to the best result for $\sigma \in [\frac{\sigma^*}{5},\infty)$, i.e., we enlarge a bit the interval of bandwidths for the homotopy. For GMS, we select $\sigma$ to give exactly the desired $K$ modes (which, as before, is cumbersome). The best results for each dataset are in boldface. $K$-modes improves over $K$-means if using a bandwidth estimated automatically, but it improves even more if searching further bandwidths. GMS gives poor results for the reasons described earlier.


\begin{table}[b]
  \begin{center}
    \caption{Clustering accuracy for high-dimensional datasets (size $N$, dimension $D$, number of classes $K$). N/A means our GMS code ran out of memory.}
    \label{t:hd-datasets}
    \begin{tabular}{@{}c|ccc|ccc@{}}
      \hline
      & \multicolumn{3}{c|}{Adjusted Rand Index (\%)} & \multicolumn{3}{c}{Normalized Mutual Info.\ (\%)} \\
      \cline{2-7}
      Dataset$(N,D,K)$ & $K$-means & $K$-modes & GMS & $K$-means & $K$-modes & GMS \\
      \hline
      COIL$(1440,1024,20)$ & 56.5 & \textbf{62.1} (62.1) & 11.6 & 76.8 & \textbf{79.1} (79.1) & 49.8 \\
      MNIST$(2000,784,10)$ & 32.9 & \textbf{35.5} (34.4) & 1.37 & 46.4 & \textbf{49.2} (47.4) & 11.7 \\
      TDT2$(9394,36771,30)$ & 55.8 & \textbf{56.1} (56.1) & N/A & 80.0 & \textbf{80.8} (80.7) & N/A \\
      \hline
    \end{tabular}
  \end{center}
\end{table}

\subsubsection*{Summary}

The previous experiments suggest that $K$-modes is more robust than $K$-means and GMS to outliers and parameter misspecification ($K$ or $\sigma$). Outliers shift centroids away from the main mass of a cluster in $K$-means or create spurious modes in GMS, but $K$-modes is attracted to a major mode within each cluster. GMS is sensitive to the choice of bandwidth, which determines the number of modes in the kde. However, $K$-modes will return exactly $K$ modes (one per cluster) no matter the value of the bandwidth, and whether the kde of the whole dataset has more or fewer than $K$ modes. $K$-means is sensitive to the choice of $K$: if it is smaller than the true number of clusters, it may place centroids in low-density regions between clusters (which are invalid patterns); if it is larger than the true number of clusters, multiple centroids will compete for a cluster and partition it, yet the resulting centroids may show no indication that this happened. With $K$-modes, if $K$ is too small the centroids will move inside the mass of each cluster and become valid patterns. If $K$ is too large, centroids from different portions of a cluster may look similar enough that their redundancy can be detected.

\section{Discussion}

While $K$-modes is a generic clustering algorithm, an important use is in appplications where one desires representative centroids in the sense of being valid patterns, typical of their cluster, as described earlier. By making $\sigma$ small enough, $K$-modes can always force the centroids to look like actual patterns in the training set (thus, by definition, valid patterns). However, an individual pattern is often noisy or idiosyncratic, and a more typical and still valid pattern should smooth out noise and idiosyncrasies---just as the idea of an ``everyman'' includes features common to most men, but does not coincide with any actual man. Thus, best results are achieved with intermediate bandwidth values: neither too large that they average widely different patterns, not too small that they average a single pattern, but just small enough that they average a local subset of patterns---where the average is weighted, as given by eq.~\eqref{e:GMS} but using points from a single cluster. Then, the bandwidth can be seen as a smoothing parameter that controls the representativeness of the centroids. Crucially, this role is separate from that of $K$, which sets the number of clusters, while in mean-shift both roles are conflated, since the bandwidth determines both the smoothing and the number of clusters.

How to determine the best bandwidth value? Intuitively, one would expect that bandwidth values that produce good densities should also give reasonable results with $K$-modes. Indeed, this was the case in our experiments using a simple bandwidth estimation rule (the average distance to the $k$th nearest neighbor). In general, what ``representative'' means depends on the application, and $K$-modes offers potential as an exploratory data analysis tool. By running the homotopy algorithm from large bandwidths to small bandwidths (where ``small'' can be taken as, say, one tenth of the result from a bandwidth estimator), the algorithm conveniently presents to the user a sequence of centroids spanning the smoothing spectrum. As mentioned before, the computational cost of this is comparable to that of running $K$-means multiple times to achieve a good optimum in the first place. Finally, in other applications, one may want to use $K$-modes as a post-processing of the $K$-means centroids to make them more representative.

\section{Conclusion and Future Work}

Our $K$-modes algorithm allows the user to work with a kernel density estimate of bandwidth $\sigma$ (like mean-shift clustering) but produce exactly $K$ clusters (like $K$-means). It finds centroids that are valid patterns and lie in high-density areas (unlike $K$-means), are representative of their cluster and neighborhood, yet they average out noise or idiosyncrasies that exist in individual data points. Computationally, it is slightly slower than $K$-means but far faster than mean-shift. Theory and experiments suggest that it may also be more robust to outliers and parameter misspecification than $K$-means and mean-shift.

Our $K$-modes algorithm can use a local bandwidth at each point rather than a global one, and non-gaussian kernels, in particular finite-support kernels (such as the Epanechnikov kernel) may lead to a faster algorithm. We are also working on $K$-modes formulations where the assignment variables are relaxed to be continuous.

A main application for $K$-modes is in clustering problems where the centroids must be interpretable as valid patterns. Beyond clustering, $K$-modes may also find application in problems where the data fall in a nonconvex low-dimensional manifold, as in finding landmarks for dimensionality reduction methods \cite{SilvaTenenb03a}, where the landmarks should lie on the data manifold; or in spectral clustering \cite{Ng_02a}, where the projection of the data on the eigenspace of the graph Laplacian defines a hypersphere.

\subsubsection*{Acknowledgments.}

Work funded in part by NSF CAREER award IIS--0754089.


\end{document}

%% file: MACPdef.tex
\newcommand{\C}{\ensuremath{\mathbf{C}}}

\newcommand{\RR}{\ensuremath{\mathbf{R}}}

\renewcommand{\c}{\ensuremath{\mathbf{c}}}

\newcommand{\f}{\ensuremath{\mathbf{f}}}

\newcommand{\x}{\ensuremath{\mathbf{x}}}

\newcommand{\bbR}{\ensuremath{\mathbb{R}}}

\newcommand{\calO}{\ensuremath{\mathcal{O}}}

\newcommand{\norm}[1]{\left\lVert#1\right\rVert}

{%
\begin{list}{#1}{
\vspace{-\topsep}
\vspace{-\partopsep}
\setlength{\itemindent}{0cm}
\setlength{\rightmargin}{0cm}
\setlength{\listparindent}{0cm}
\settowidth{\labelwidth}{#1}
\setlength{\leftmargin}{\labelwidth}
\addtolength{\leftmargin}{\labelsep}
\setlength{\itemsep}{0cm}
}%
}%
{%
\end{list}
\vspace{-\topsep}
\vspace{-\partopsep}
}

{\begin{enumerate}%
}%
{\end{enumerate}}

%% file: MACPdef-ams.tex
\theoremstyle{plain}

\newtheorem*{lemma*}{Lemma}

\newtheorem*{prop*}{Proposition}

\theoremstyle{definition}

\newtheorem*{defn*}{Definition}

\newtheorem*{exmp*}{Example}

\newtheorem*{conj*}{Conjecture}

\theoremstyle{remark}

\newtheorem*{rmk*}{Remark}
